\documentclass[journal,twoside,web]{ieeecolor}
\usepackage{jsen}
\usepackage{cite}
\usepackage{amsmath,amssymb,amsfonts}
\usepackage{graphicx}
\usepackage{textcomp}
\usepackage{wrapfig}
\usepackage{svg}
\usepackage{makecell}

\def\BibTeX{{\rm B\kern-.05em{\sc i\kern-.025em b}\kern-.08em
    T\kern-.1667em\lower.7ex\hbox{E}\kern-.125emX}}
\definecolor{abstractbg}{rgb}{0.89804,0.94510,0.83137}
\usepackage{booktabs} % For formal tables
\usepackage{ifpdf}
\usepackage{url}
\usepackage{xspace}
\usepackage{balance}
\usepackage{caption}
\usepackage[tight,footnotesize]{subfigure}
\usepackage{multirow}
\usepackage{color}
\usepackage{comment}
\usepackage{algorithm}
\usepackage{algpseudocode}
\usepackage{tabularx}

\floatname{algorithm}{Algorithm}

\usepackage{makecell} % For newline in tables

\usepackage{mathtools}

\newcommand{\etal}{\textit{et al}.\xspace}
\newcommand{\ie}{\textit{i}.\textit{e}.\xspace}

\newcommand{\update}[1]{\textcolor{black}{#1}}

\begin{document}

\title{SitPose: Real-Time Detection of Sitting Posture and Sedentary Behavior Using Ensemble Learning With Depth Sensor}
\author{Hang~Jin,
Xin~He,~\IEEEmembership{Member,~IEEE},
	Lingyun~Wang,
	Yujun~Zhu,
	Weiwei~Jiang,
    and~Xiaobo~Zhou,~\IEEEmembership{Senior~Member,~IEEE}
\thanks{This work has been in part supported by the Natural Science Foundation of China under grant No. 62072004. Corresponding author: {\it Xin~He, Yujun~Zhu.}}
\thanks{X. He, H. Jin, L. Wang, and Y. Zhu are with the School of Computer and Information, Anhui Normal University, WuHu, 241002, Anhui, China (e-mail: \{xin.he, lingyunwang, zhuyujun, hang.jin\}@ahnu.edu.cn).}
\thanks{W. Jiang is with the School of Computer Science, Nanjing University of Information Science and Technology, 210044, Jiangsu, China (e-mail: weiwei.jiang@nuist.edu.cn). }
\thanks{X. Zhou is with the School of Computer Science and Technology, Tianjin University, Tianjin, China (email: xiaobo.zhou@tju.edu.cn).}
}

\IEEEtitleabstractindextext{%
\fcolorbox{abstractbg}{abstractbg}{%
\begin{minipage}{\textwidth}%
\begin{wrapfigure}[12]{r}{3in}%
\includegraphics[width=2.75in]{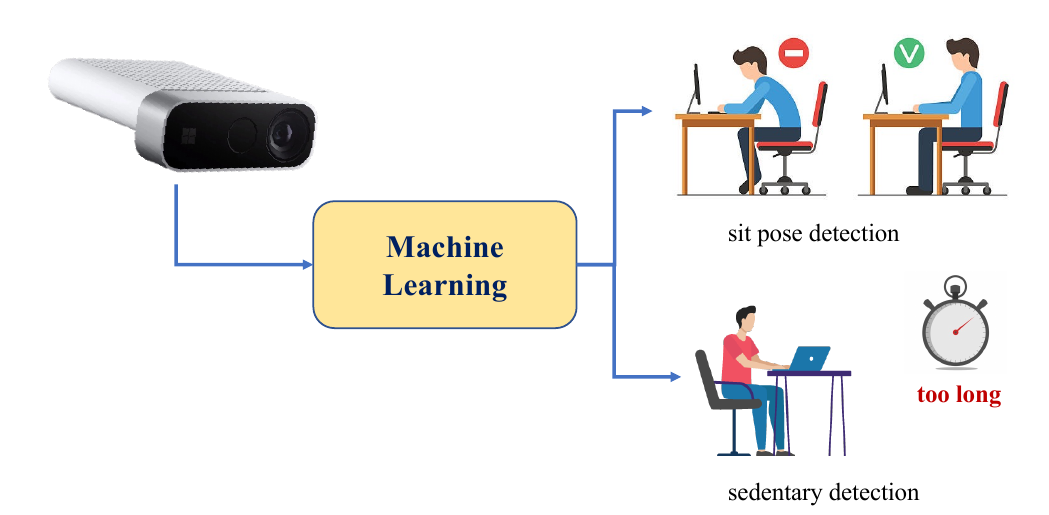}%
\end{wrapfigure}%
\begin{abstract}
Poor sitting posture can lead to various work-related musculoskeletal disorders (WMSDs). Office employees spend approximately 81.8\% of their working time seated, and sedentary behavior can result in chronic diseases such as cervical spondylosis and cardiovascular diseases. To address these health concerns, we present SitPose, a sitting posture and sedentary detection system utilizing the latest Kinect depth camera. The system tracks 3D coordinates of bone joint points in real-time and calculates the angle values of related joints. We established a dataset containing six different sitting postures and one standing posture, totaling \update{33,409} data points, by recruiting \update{36} participants. We applied \update{several state-of-the-art} machine learning algorithms to the dataset and compared their performance in recognizing the sitting poses. Our results show that the ensemble learning model based on the soft voting mechanism achieves the highest F1 score of \update{98.1}\%. Finally, we deployed the SitPose system based on this ensemble model to encourage better sitting posture and \update{to reduce} sedentary habits.
\end{abstract}

\end{minipage}}}

\maketitle

\section{Introduction}  

Office workers typically remain seated throughout their workday due to the nature of their tasks and various other factors. Consequently, many experience backaches, primarily due to their poor sitting posture and prolonged sedentary habits. Medical research indicates that improper sitting posture can lead to a range of health issues, particularly affecting the cervical and lumbar spine, and significantly \update{affecting respiratory function}~\cite{szczygiel2017musculo}. Furthermore, prolonged sitting can double the risk of developing diabetes, as well as contribute to the accumulation of abdominal fat, leading to health problems such as overweight, obesity, and cardiovascular disease~\cite{Jang2020Comprehensive}.

In the field of occupational health research, Kallings \etal~\cite{kallings2021workplace} conducted a comprehensive study demonstrating a correlation between increased durations of sedentary behavior in the workplace and a decline in self-reported general health status. This extensive study encompassed a sample size of 44,978 individuals, with an average participant age of 42.1 years, ranging from 18 to 75 years. Concurrently, Cao~\etal~\cite{cao2022associations}. provided insights into the health implications of prolonged sedentary lifestyles. Their research, involving a substantial cohort of 360,047 participants from the UK Biobank, delved into the relationship between sedentary behavior (exceeding 6 hours per day) and the heightened risk of 12 types of non-communicable diseases (NCDs).

Recognizing these significant health implications, our research aims to mitigate such risks by introducing a novel sitting posture health detection system that utilizes visual detection technology to provide interactive reminders. This system improves the sitting posture and habits of individuals, thereby preventing chronic diseases caused by prolonged sitting, especially in working scenarios. The RoSeFi~\cite{RoSeFi} system adopted WiFi channel state information to monitor sedentary behavior. However, we also aim to examine sitting poses in real-time. Therefore, we have chosen a depth camera as the monitoring sensor.

In 2019, Microsoft released Azure Kinect DK (Developer Kit), which is the third Kinect depth camera after Kinect V1 and Kinect V2. Unlike the previous Kinect V1 and Kinect V2, which are mainly aimed at Xbox game consumers, Azure Kinect is a platform for professional developers and commercial companies. It features advanced AI sensors and accompanying computer vision and speech model development kits. The Azure Kinect camera group consists of an RGB camera and an infrared camera. It continues the time-of-flight (ToF) depth estimation principle of Kinect V2, which calculates the distance to the target by calculating the time required for the emitted light to reach the target and return to the sensor \cite{tolgyessy2021evaluation}. Its accuracy is significantly higher than that of other commercial cameras. The latest Kinect camera excels not only in terms of vision but also in hearing and motion perception. It has an array of seven microphones for sensing sound. Kinect also integrates inertial sensors for various types of perception. It is worth mentioning that Microsoft provides an AI algorithm toolkit based on deep learning and convolutional neural networks (CNN) and develops a set of human body tracking SDK \cite{tolgyessy2021skeleton}.

Ran~\etal~\cite{ran2021portable} developed and designed a pressure sensor-based sitting posture detection system and used machine learning for classification. Guduru \etal~\cite{guduru2022prediction} collected RGB images of upper body poses, preprocessed them with bandpass filters, and finally used deep learning to identify posture changes during sedentary work. Roh \etal~\cite{roh2018sitting} deployed pressure sensors on the back of the chair cushion along with the backrest and implemented a sitting posture classification system using a support vector machine (SVM). Existing RGB cameras and pressure sensor seats have limitations in detecting sitting posture and sedentary status. 

The Azure Kinect depth camera offers several advantages over traditional RGB cameras, particularly in the context of posture monitoring. Unlike RGB cameras, which have stringent requirements for light sources and are associated with high costs and complexity in sensor-based deployment, the depth camera operates effectively even in sub-optimal lighting conditions. This feature not only facilitates easier monitoring of sitting postures but also ensures user privacy is preserved. Furthermore, the ease of deployment of the camera eliminates the need for complex operations like installing pressure sensors on chairs, making it a more convenient and privacy-conscious choice for monitoring sitting states.

Bontrup~\etal~\cite{bontrup2019low} recruited three trial participants and used the Kinect to collect depth data. Machine learning was utilized to categorize sitting postures into two types: correct and incorrect postures. Paliyawan~\etal~\cite{paliyawan2014prolonged} employed Kinect to gather time and depth coordinates for ten key joints, using machine learning to help office workers avoid the phenomenon of prolonged sitting. Compared to the state-of-the-art studies, SitPose collects posture data from \update{36} participants and is capable of identifying two harmful behaviors, \ie, prolonged sitting and incorrect sitting postures. The real-time system promptly reminds users to improve their posture through pop-up notifications. Additionally, we categorized the types of incorrect postures, enabling users to understand their habitual incorrect sitting positions. This knowledge empowers users to avoid such postures in the future, contributing to better ergonomic health.

The depth camera works based on the principle of TOF technology. Specifically, when the near-infrared light emitted by the sensor encounters a human body or an object, it will be reflected. The distance between the human body or the object can be calculated by computing the difference between the emitted light time and the reflected time or the phase difference, and depth information can be obtained. Then, the human body tracking software development kit (SDK) provided by Kinect can be used to identify human joints and bones.  

In this paper, we design a sitting posture recognition system, namely {\it SitPose}, which classifies the sitting postures from the captured three-dimensional bone data using machine learning. The initial step of the system involves utilizing Azure Kinect as a monitoring tool, which optimizes the data collected from depth image through feature extraction. SitPose also calculates the angles of human bone joints as part of the features. Finally, it employs ensemble learning to classify different sitting postures. To evaluate its classification effect, we use confusion matrices and conduct abnormal experiments. Based on the information of TOF, Kinect's recognition principle, and a soft voting mechanism in ensemble learning, we successfully built a sitting posture recognition system.

The main contributions of this paper are summarized as follows. 
\begin{itemize}
    \item We design a real-time sitting posture monitoring system using a depth camera. The latest Azure Kinect is utilized for processing photos and presenting bone joint points directly, which offers ease of deployment and addresses privacy concerns. %We apply the SVM classifier for classification.
    \item For evaluating the performance, we generate a dataset of \update{36} participants. The dataset has a total of \update{33,409} samples, which is a large number. It contains body depth values of skeletal coordinates. \update{We employed $k$-fold cross-validation to train our model, which offers a more comprehensive evaluation of the model's accuracy and efficiency. To visualize the results more directly, we applied a confusion matrix. Using 5-fold cross-validation, the training set is divided into five parts. In each iteration, four parts are used for training and one part for testing, repeating this process five times. Finally, we calculate the average of the results from these five test iterations.}
    \item To recognize sitting postures, we adopted an ensemble learning approach based on a \update{soft voting mechanism, SVM, decision trees (DT),  multilayer perceptron (MLP), gradient boosting decision trees (GBDT) and TabNet}. Comprehensive experiments were conducted to evaluate the recognition accuracy of the respective models. Additionally, we demonstrated the system's latency to showcase its real-time capability.
\end{itemize}

Therefore, the system has a particular auxiliary effect on improving sitting posture and preventing lousy sitting posture and some chronic diseases caused by sedentary sitting.

The rest of this paper is organized as follows. \update{Section II introduces the depth estimation principle of ToF and how Kinect realizes the recognition of human bone joints.} In section III, we discuss the approach of utilizing Azure Kinect as a monitoring tool as well as incorporating calculated joint angles into the feature set. Section IV presents a comparison of the classification performance of \update{six} different algorithms. In this case, we analyze the results of each action through a confusion matrix. In Section V, we summarize the effect of the detection system and propose what can be improved in future research.

\section{Preliminary Knowledge}
In this section, we introduce the depth estimation principle of ToF, how Kinect realizes the recognition of human bone joints, and the principle of the soft voting mechanism in ensemble learning.

\subsection{Time of Flight}
The previous generation Kinect V2, as well as the Azure Kinect, both employ time-of-flight technology. The basic principle of ToF is to calculate the transmission time of a signal based on its reflection to determine the distance from a person or an object \cite{castaneda2011time}. Specifically, a depth camera emits a continuous pulse of infrared light, which would be reflected back when it encounters a human body or an object. By recording the time-of-flight from sending the pulse to receiving the pulse and using continuous wave modulation and phase detection inside the camera, the distance of the human body or object can be calculated \cite{YuGAO2022Kinect}. The modulation measurement principle of the continuous sine wave is as follows: $S(t)$ is a sinusoidal signal sent by the depth camera, $a$ is the amplitude, the modulation frequency is $f$, and $S(t)$ can be expressed as
\begin{equation}
S(t)=a(1+\sin(2\pi ft)).
\end{equation}

\begin{figure}[t]
    \centering
    \includegraphics[width=2in]{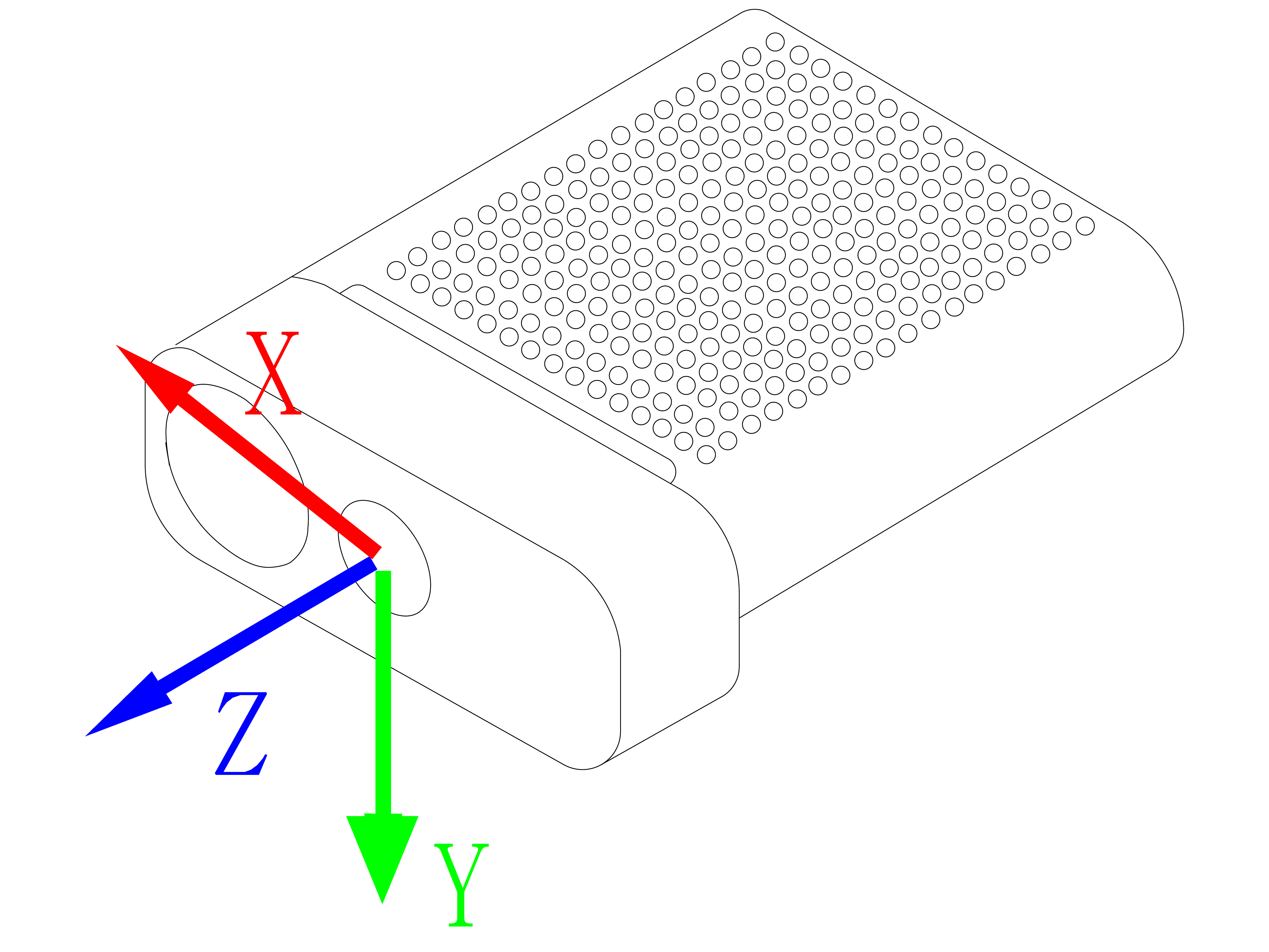}
    \caption{Azure Kinect's spatial coordinate axis.}
    \label{fig:coordinate}
\end{figure}

The signal received after being reflected by the object is $R(t)$, specifically
\begin{equation}
R(t)=A(1+\sin(2\pi ft-\Delta\varphi))+B,
\end{equation}
where $A$ represents the amplitude of the attenuated signal after reflection, $B$ represents the intensity shift caused by ambient light, and $\Delta\varphi$ represents the phase difference. To set these parameters, $R(t)$ is sampled 4 times at intervals of $\frac{T}{4}$, and a system of equations containing 4 equations is constructed using the 4 sampling values, as
\begin{multline}
R_i=A\sin(2\pi f(t_i-\Delta t))+(A+B),
\\i=0,1,2,3
\end{multline}
where $\Delta t$ represents the delay between the time when the sinusoidal signal is emitted and the time when the reflected signal is received. The phase difference $\Delta\varphi$ can be obtained by constructing a system of equations using the four samples, as
\begin{equation}
\Delta\varphi=\arctan2(R_2-R_0,R_1-R_3).
\end{equation}

Finally, based on the calculated phase difference, the depth distance $d$ is obtained as
\begin{equation}
d=\frac{c}{4\pi f}\Delta\varphi,
\end{equation}
where $c$ is the speed of light.

\subsection{Human Skeleton Joints}
The detection and recognition of human skeleton joints has a wide range of applications in motion and posture perception. Azure Kinect Sensor SDK initially processes the infrared (IR) image through a CNN. This step is crucial for identifying the positions of 2D joints and segmenting various body parts within the image. Following this information, the depth image, in conjunction with the previously determined 2D joint locations, is integrated into a skeleton-based tracking model. This sophisticated approach yields several key outputs: the precise locations of 3D joints, their respective orientations, and the tracking of their movements over time \cite{liu20193d, Microsoft1, Microsoft2}. Microsoft has also developed a new body-tracking SDK for the Azure Kinect, which is based on Deep Learning and CNN. However, the Azure Kinect Body Tracking SDK is closed-source. The Azure Body Tracking SDK, utilizing the power of AI, intelligently estimates the coordinates of obscured extremities. This is achieved through the application of decision forests, which have been trained on an extensive dataset \cite{jo2022agreement}.
We can utilize the API interface provided by Microsoft to capture depth data for human body joint points.
\begin{table}[t]
    \centering
    \caption{{Notations.}}
    \begin{tabularx}{3in}{cc}
        \toprule
        \textbf{Notation} & \textbf{Meaning} \\ \midrule
        % $w$ & \makecell{normal vector that \\ determines the direction of the hyperplane}\\ 
        % $b$ & \makecell{distance between \\ the hyperplane and the origin} \\ 
        % $x$ & a point in the sample space \\ 
        $S(t)$ & a sinusoidal signal sent by the depth camera \\ 
        $a$ & transmitted amplitude \\ 
        $f$ & modulation frequency \\ 
        $A$ & attenuated amplitude at the receiver \\
        $B$ & intensity shift caused by ambient light \\ 
        $\Delta \varphi$ & phase difference\\ 
        $R(t)$ & signal received at time $t$\\ 
        $\Delta t$ & delay\\ 
        $d$ & depth distance\\ 
        $c$ & the speed of light\\ \bottomrule
    \end{tabularx}
    \label{tab:notation}
\end{table}

As shown in Fig.~\ref{fig:coordinate}, the Kinect skeletal joint coordinate system takes the depth camera focus as the coordinate origin, which is at $[0,0,0]$. The positive X-axis extends to the right along the depth camera focus, the positive Y-axis is vertically downward, and the Z-axis extends directly in front of the camera focus. 

\subsection{Voting Classifier}
\label{AA}
To enhance the accuracy of sitting posture detection and mitigate the issue of poor generalization when using a single model, we propose an ensemble learning model based on a voting classifier. Rather than being a standalone machine learning algorithm, this approach combines multiple machine learning models to complete the learning task. The voting classifier works by aggregating the predictions of several models through a voting mechanism, thereby reducing variance and improving the robustness of the model.

There are two types of voting methods: hard voting and soft voting. The key difference between the two is that hard voting predicts the class with the most votes, while soft voting predicts the class with the highest sum of predicted probabilities from all models. In other words, hard voting relies on the direct class label outputs of each model, whereas soft voting utilizes the probability outputs from each model to make predictions.

\begin{figure}[t]
    \centering
    \includegraphics[width=3.5in]{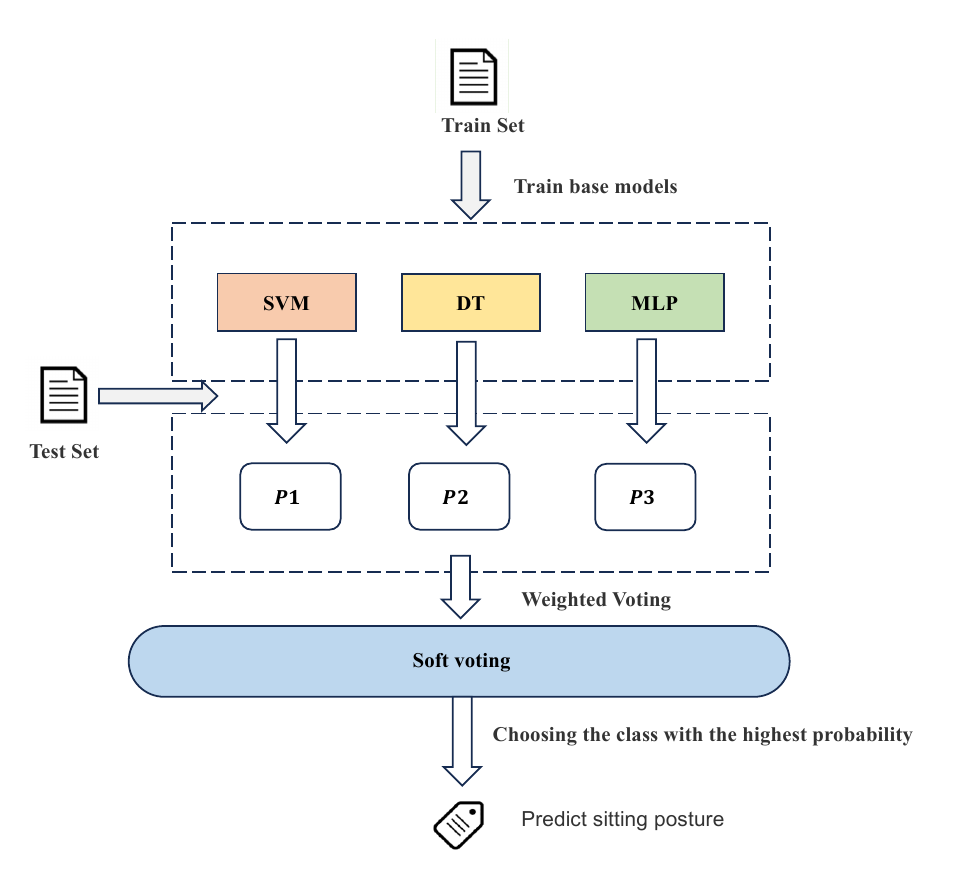}
    \caption{The framework of the sitting posture and sedentary habits detection system.}
    \label{fig:flowchart}
\end{figure}

As shown in Fig.~\ref{fig:flowchart}, this paper employs three well-established models: SVM, DT, and MLP. These models are chosen because they use different learning strategies and learn features from different perspectives and spaces, complementing each other to enhance the overall performance of the voting ensemble learning model. For the voting method, we opted for soft voting, which uses the predicted probabilities from each model. This approach suits the three selected machine learning models more than hard voting.

%%%%%%%%%%%%%%%%%%%%%%%%%%%%%%%%%%%%%%%%%%
\section{System Design}
In this section, we introduce the method of adopting Azure Kinect to monitor sitting posture detection and the functional modules of the detection system. The adopted notations are summarized in Table~\ref{tab:notation}. The detection system is based on the human skeleton coordinates collected by the depth image. It first optimizes the data through feature extraction, then calculates the angles of relevant bone joints and adds them to the features. Ensemble learning models are used to classify different sitting postures. Machine learning models are used to classify different sitting postures. Finally, we deploy the trained model, take real-time depth images from Kinect, and input bone data into the model for classification.

Seven postures are classified as below: 
\begin{itemize}
    \item[-] sitting straight
    \item[-] hunching over
    \item[-] left sitting
    \item[-] right sitting
    \item[-] leaning forward
    \item[-] lying
    \item[-] standing
\end{itemize}
% Seven postures are classified as sitting straight, hunching over, left sitting, right sitting, leaning forward, lying and standing.

\begin{figure}[t]
    \centering
\includegraphics[width=3in]{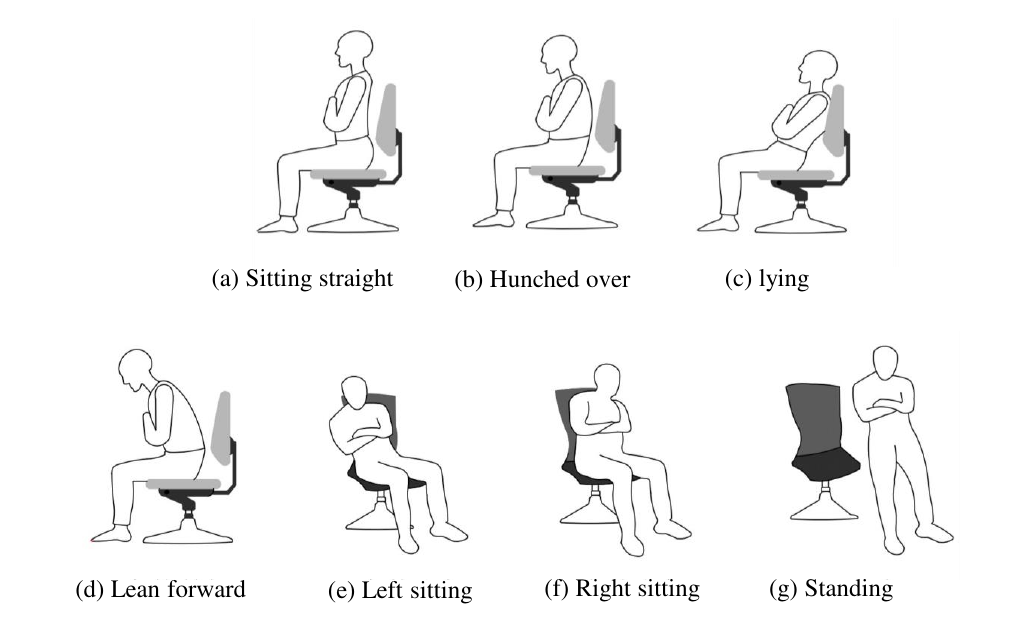}
    \caption{The illustration of six types of sitting posture and standing which we examine in this work.}
    \label{fig:types}
\end{figure}

These six chosen sitting postures are selected based on comprehensive studies on sitting positions~\cite{zemp2015sitting, GRANDJEAN1977135}. \update{These postures represent a broad spectrum of typical office postures, as shown in} Fig.~\ref{fig:types}. We include a standing event as a detection feature to determine if the user is seated on an office chair. This addition significantly enhances the accuracy of identifying sedentary behaviors.

\subsection{System Overview}
As shown in Fig.~\ref{fig:flowchart}, we present a posture detection framework based on heterogeneous ensemble learning techniques, specifically utilizing a soft voting mechanism to enhance prediction accuracy and robustness\footnote{The dataset and code are available on \textcolor{blue}{\url{https://github.com/TaylorDurden1114/SitPose}}.}. The framework comprises three core modules: data collection and preprocessing, soft voting ensemble learning, and prediction.

First, the data collection and preprocessing module gathers depth camera data of specified skeletons using the Azure Kinect and extracts critical features such as skeletal angles.

The core of the framework is the soft voting ensemble learning module, which combines predictions from various algorithms, including DT, SVM, and MLP. This ensemble approach effectively integrates the strengths of individual models by averaging their weighted predictions, thereby improving overall performance and reliability compared to traditional single models.

Finally, the prediction module uses the output from the ensemble learning module to make the final posture assessment, providing real-time feedback and recommendations to help users improve their posture and reduce health issues caused by prolonged poor sitting habits. The framework aims to deliver a comprehensive and accurate solution for posture detection through efficient data processing, advanced ensemble learning techniques, and precise prediction capabilities.

\subsection{Collection of Joint Coordinate Data}
\update{Although} many studies have made Kinect-related datasets \update{publicly} available, there \update{ is still a noteble lack of depth data specifically dedicated to sitting postures.}  \update{To address this gap, we} gathered sitting posture data from \update{36} participants, \update{ensuring an identical} gender ratio, and \update{gathered a total of 33,409} instances of raw depth data. Table~\ref{tab:sample} presents information about participants. This dataset \update{aims to facilitate} future \update{research by allowing additional features extraction} from the raw data for posture analysis purposes, \update{thereby advancing} the field. 

\begin{figure*}[t]
\centering
\begin{tabular}{cccc}
\includegraphics[width=0.22\textwidth]{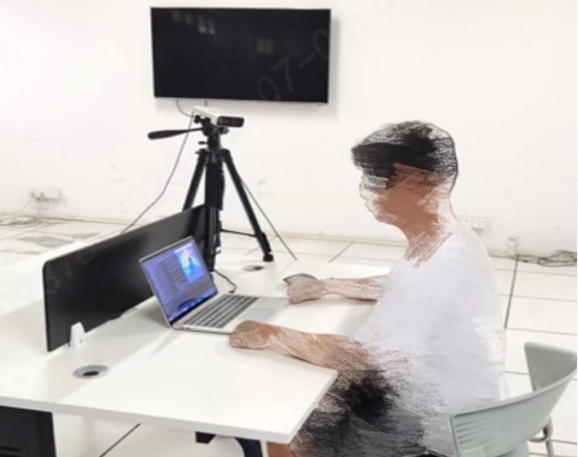} &
\includegraphics[width=0.22\textwidth]{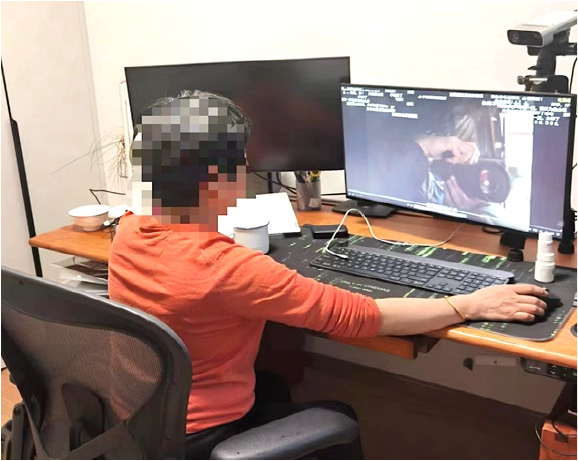} & 
\includegraphics[width=0.22\textwidth]{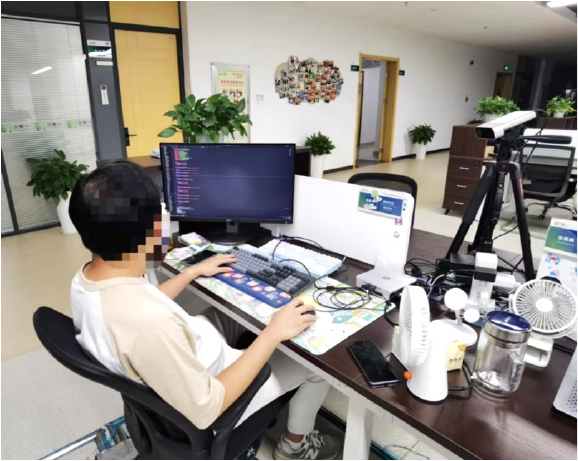} &
\includegraphics[width=0.22\textwidth]{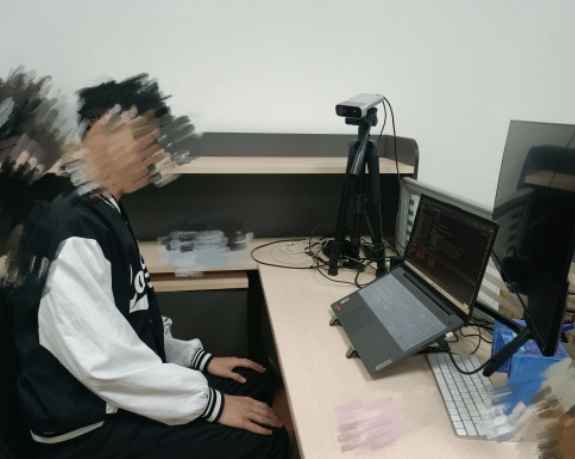} \\
(a) hardware lab & (b) home & (c) company & (d) office \\ [6pt]
\end{tabular}
\caption{Experimental environments in our study.}
\label{fig:environment}
\end{figure*}

We installed two development kits, Azure Kinect sensor SDK and Body Tracking SDK, on a Win10 system computer. Table~\ref{tab:setup} \update{outlines} the components and specifications of the experimental setup. The experimental \update{environments are illustrated} in Fig.~\ref{fig:environment}. \update{Experiments were conducted in four different settings: a hardware lab, a home, a company office, and an office at our university. Additionally, the depth camera was positioned on both the right and left sides of the table to provide flexibility for real-world deployment scenarios. Participants performed six sitting postures (sitting straight, hunching over, leaning left, leaning right, leaning forward, and reclining) as well as a standing posture in front of the table.} By \update{collecting sitting posture data from various social roles and environments}, we aim to ensure a certain level of generalizability for the model.

\update{While the Azure Kinect can capture 32 skeletal points, this study focuses on only 9 key skeletal points for several reasons. In a desk-based office setting, leg joints are frequently obstructed by the desk, and although algorithms can estimate leg positions, the accuracy of such data is limited. Hand joints were further excluded due to potential inaccuracies caused by frequent hand movements during office activities, e.g., drinking, writing, using a mouse, or browsing documents. Furthermore, certain head joints—such as the nose, left eye, and right eye—show minimal positional variation, making them redundant for our purposes. Therefore, only 9 key skeletal points were selected for data collection based on these considerations}.

\begin{table}[t]
\centering
\caption{\update{Participant summary statistics for height, weight, and age, categorized by gender. The table displays the mean and standard deviation (Mean ± Std), minimum, and maximum values for both male and female participants, as well as the overall group. }}
\begin{tabular}{l l c  c c}
  \toprule
\textbf{Parameter} & \textbf{Gender (count)} & \textbf{Mean} $\pm$ \textbf{Std} & \textbf{Min} & \textbf{Max} \\
\hline
\textbf{Height (cm)}   & Male (18)  & 174.28  $\pm$ 4.82  & 167 & 185 \\
                   & Female (18) & 162.06 $\pm$ 4.49 & 155 & 171 \\
                   & Overall (36) & 168.17  $\pm$ 7.71 & 155 & 185 \\ \hline
\textbf{Weight (kg)}  & Male (18)  & 68.36   $\pm$  8.74  & 50  & 87  \\
                   & Female (18) & 53.39  $\pm$  6.02  & 45  & 66  \\
                   & Overall (36) & 60.88  $\pm$ 10.6  & 45  & 87  \\ \hline
\textbf{Age}     & Male  (18)  &   23.67    $\pm$  2.95    & 20    & 29    \\
                   & Female (18)  &  25.83     $\pm$  10.12   & 20    & 57    \\
                   & Overall (36) &    24.75   $\pm$ 7.43    & 20    & 57    \\
  \bottomrule
\end{tabular}
\label{tab:sample}
\end{table}

Through Kinect's software development kit, we can obtain human body depth images and bone joint data in real-time. Each joint has a unique index value, which allows us to obtain specific joint data \cite{Yun2021Review}. Azure Kinect can track and accurately extract 32 bone coordinates of the human body in real-time, which is 7 more than Kinect V2. We select \update{nine} joints of the subject to identify per second while also saving the coordinates of them; these nine joints are the subject's head, left clavicle, right clavicle, neck, right shoulder, left shoulder, spine of the chest, spine of the navel, and pelvis. Therefore, each sampling will obtain the raw data of $1\times24$ joint coordinate data. The height of the experimenters ranged from 155 to 185 cm, and experimenters of different heights were tested to avoid systematic classification errors caused by height differences. In addition, during the experiment, we slightly changed the seat position and allowed the experimenter to adjust the range of sitting posture movements to enhance the robustness of the sitting posture detection system and make the system deployment more flexible.

\subsection{Joint Angle Feature Extraction}
\begin{figure}[t]
    \centering
    \includegraphics[width=3in]{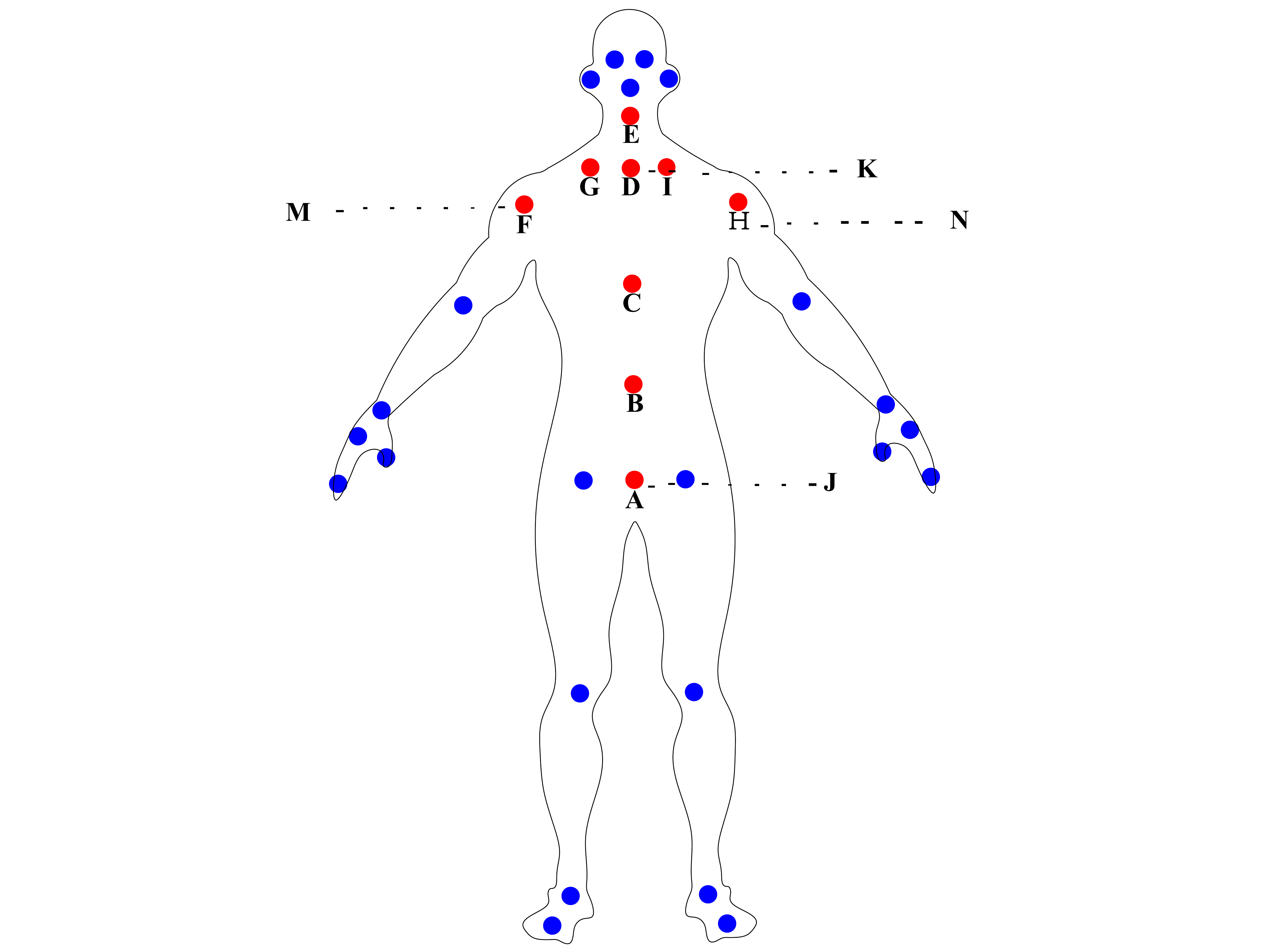}
    \caption{The recognized joints by the depth camera where the selected joints are marked in red.}
    \label{fig:feature}
\end{figure}

The angles between the bones and joints of different sitting postures are various, and these angles are an important factor in distinguishing sitting postures. To this end, we can calculate the spatial angle between the relevant bone joints, add it to the features of the data set, and use the angle to distinguish the sitting posture. We consider the angle information between joints an important parameter for training the model. First, the angle calculation is performed on the three-dimensional bone coordinates, and the angle data related to the three-dimensional coordinates of the joints of the neck, head, left shoulder, right shoulder, chest, and pelvis are obtained. Here, taking the joint angle between the thoracic spine, cervical spine, and head
\begin{math}
(\overrightarrow{ED},\overrightarrow{DC})
\end{math}
as an example, the calculation steps of the relevant angles are introduced.
Assuming that the three-dimensional coordinates of the head are 
\begin{math}
    p_1=(x_1,y_1,z_1)
\end{math}
and the three-dimensional coordinates of the neck are calculated by
\begin{math}
    p_2=(x_2,y_2,z_2)
\end{math}
then the spatial vector from the neck to the head 
\begin{math}
   \overrightarrow{P_1P_2} 
\end{math}
is: 
\begin{equation}
    \overrightarrow{P_1P_2}=(x_2-x_1,y_2-y_1,z_2-z_1)
\end{equation}
Further, we assume that the three-dimensional coordinates of the chest are
\begin{math}
    p_3=(x_3,y_3,z_3)
\end{math}
then the spatial vector
\begin{math}
   \overrightarrow{P_2P_3}
\end{math}
from the neck to the chest is: 
\begin{equation}
    \overrightarrow{P_2P_3}=(x_3-x_2,y_3-y_2,z_3-z_2).
\end{equation}
Let $|\cdot|$ stand for the norm of a vector. Due to two vectors and their norm are known, the angle $\theta$ between the spatial vectors 
\begin{math}
    \overrightarrow{P_1P_2},
    \overrightarrow{P_2P_3} 
\end{math}
is calculated as
\begin{equation}
    \theta=\arccos(\frac{\overrightarrow{P_1P_2}\cdot \overrightarrow{P_2P_3}}{\lvert \overrightarrow{P_1P_2}  \rvert \cdot \lvert \overrightarrow{P_2P_3}  \rvert})\times \frac{180^\circ}{\pi}.
\end{equation}
To further enhance the accuracy and precision of sitting posture analysis, we introduced several auxiliary points based on skeletal points to assist in measuring and describing the angles between key skeletal points. These auxiliary points include point $J$, parallel to the x-axis of the pelvis (point A), point K, parallel to the x-axis of the neck (point D), and points M and N, parallel to the x-axis of the left clavicle (point G) and right clavicle (point H), respectively. For example, the spatial angle $\angle JAB$ was calculated from the pelvis to the abdomen, which reflects the degree of spinal curvature in different sitting postures. The letters marked by the human body joints are shown in Fig.~\ref{fig:feature}, after experiments we select 
\begin{math}
    (\overrightarrow{JA},\overrightarrow{AB}),
    (\overrightarrow{JA},\overrightarrow{AC}),
    (\overrightarrow{JA},\overrightarrow{AD}),
    (\overrightarrow{AB},\overrightarrow{BC}),
    (\overrightarrow{BC},\overrightarrow{CE}), \\
    (\overrightarrow{NI},\overrightarrow{IH}),
    (\overrightarrow{MG},\overrightarrow{GF}),
    (\overrightarrow{HI},\overrightarrow{IC}),
    (\overrightarrow{FG},\overrightarrow{GC}).
\end{math}
The corresponding angles between these nine joint spatical vectors have been included to the feature for further processing.  

\begin{figure*}[t]
\centering
\begin{tabular}{ccc}
\hspace{-0.7cm}\includegraphics[width=0.35\textwidth]{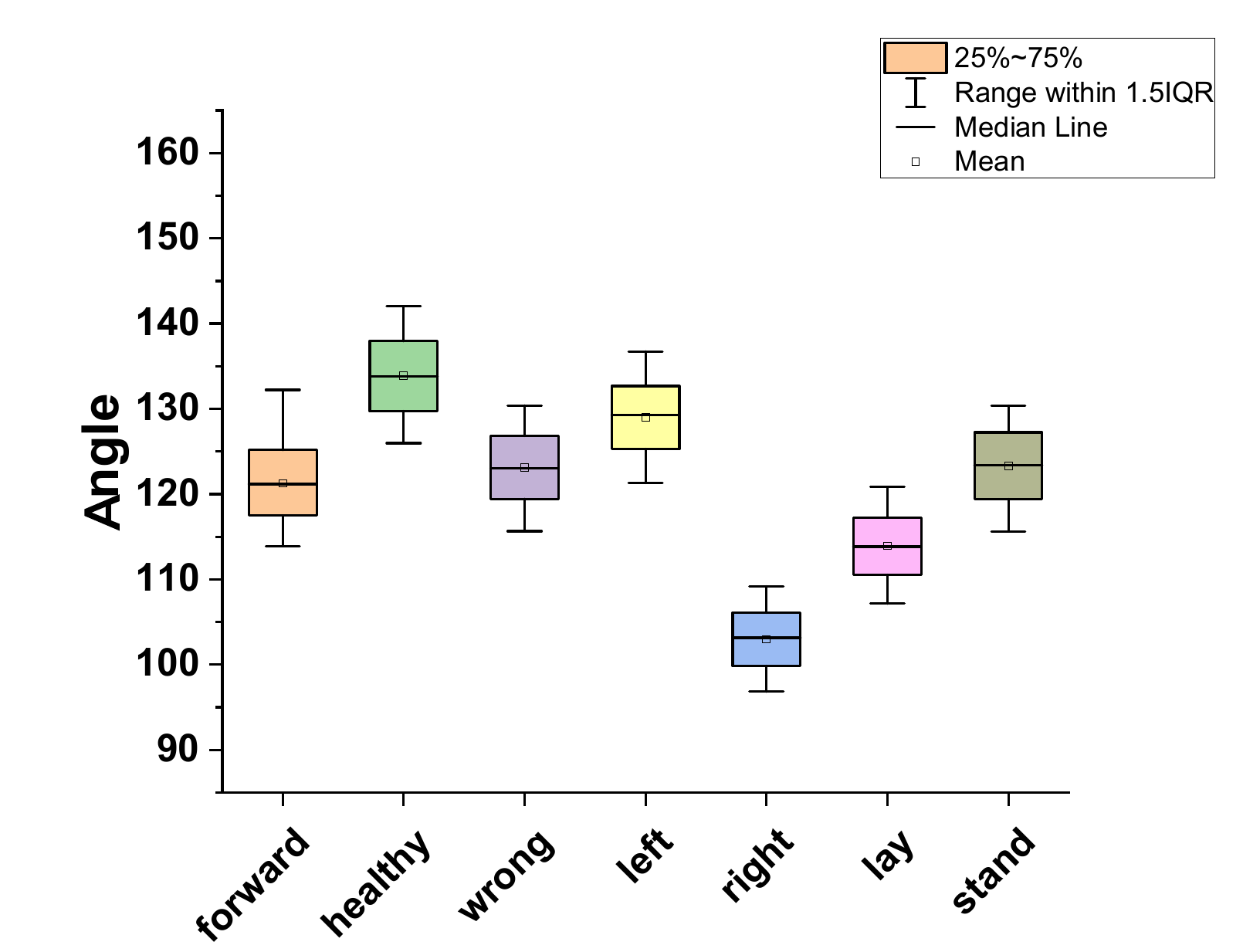} &
\hspace{-0.7cm}\includegraphics[width=0.35\textwidth]{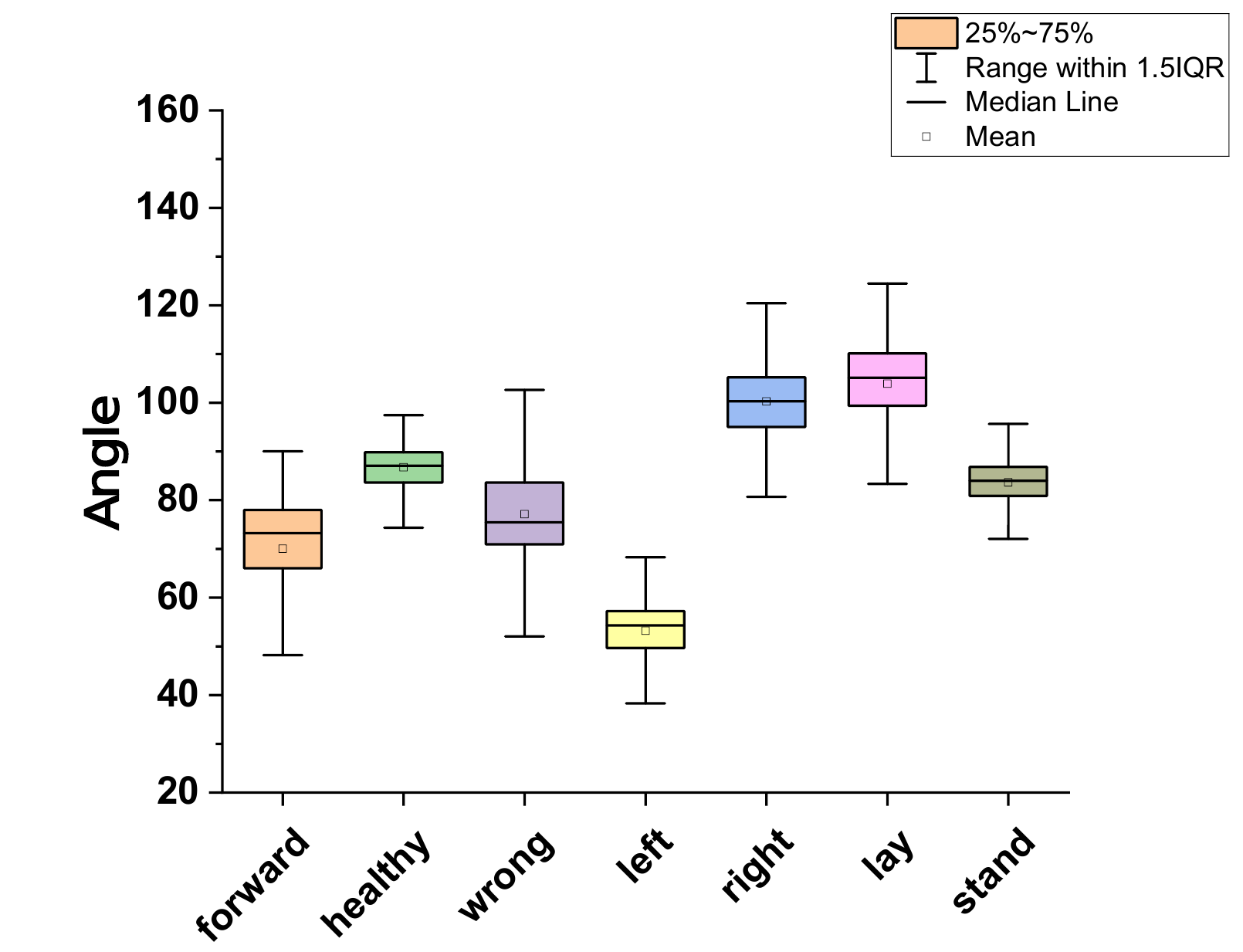} & 
\hspace{-0.7cm}\includegraphics[width=0.35\textwidth]{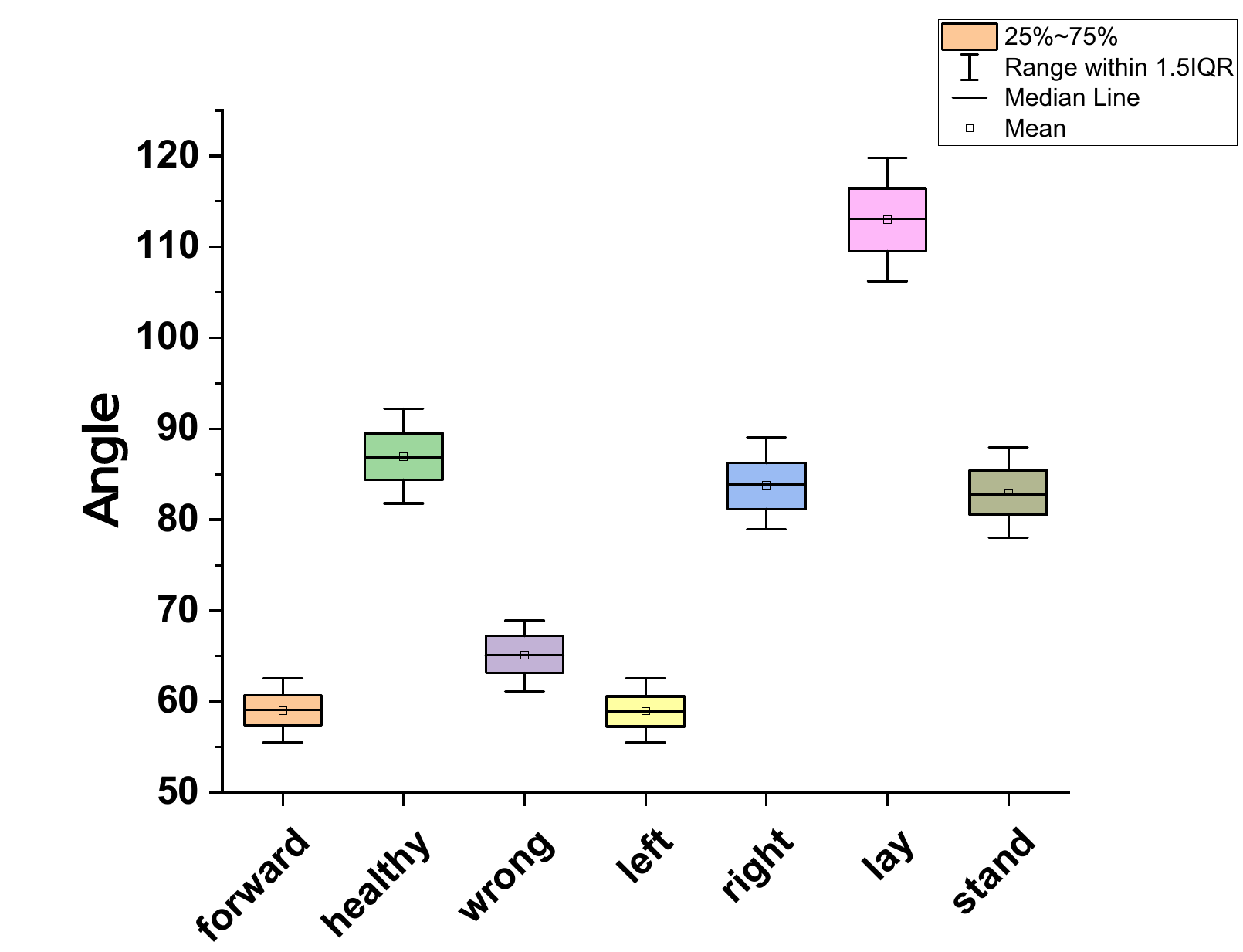} \\
(a) BCE & (b) JAB & (c) KDE \\ [6pt]
\end{tabular}
\caption{Illustration the distribution of skeletal angle information across different postures for three methods: (a) Points B-C-E (BCE), (b) Points J-A-B (JAB), and (c) Points K-D-E (KDE).}
\label{fig:Afeature}
\end{figure*}

We conducted a preliminary study to show the variations in the angles of our selected skeletal points, where the results are shown in Fig.~\ref{fig:Afeature}. From the results, the distribution of skeletal angle information can be demonstrated. For instance, the data distribution from the joint points KDE shows that the leaning-back posture can be clearly distinguished from other sitting postures. In contrast, healthy sitting, right-leaning sitting, and standing postures can be differentiated from forward-leaning, hunchback, and left-leaning sitting postures. This indicates that different sitting postures exhibit significant differences in the distribution characteristics of skeletal angles, which helps in more accurate posture classification and assessment analysis.

\subsection{Composition of the Dataset} 
The sitting posture detection data set has a total of \update{33,409} items, including sitting straight, hunched over, left sitting, right sitting, leaning forward, lying, and standing. Among them, \update{5,299} are sitting straight, \update{4,985} are hunched over, \update{4,715} are leaning forward, \update{4,259} are left sitting, \update{4,337} are right sitting, \update{4,340} are lying and \update{5,474} are standing.

By analyzing depth images and depth values, we found that the depth data of the head is very useful for distinguishing between left sitting and right sitting postures. Therefore, we included the head`s depth data as a feature in the dataset.
The dataset comprises the head`s depth values and nine extracted key skeletal angle features.

To ensure unbiased and broader model performance, we utilized $k$-fold cross-validation to evaluate the model, aggregating the evaluation metrics through averaging.
The parameters of the model are trained based on the data of the training set, and the performance of the trained model is evaluated using the test set. 
We evaluated the performance of our collected human sitting posture dataset using an ensemble learning model with a soft voting mechanism.

\subsection{Parameter Settings of Classification Models}
We adopted SVM, DT, and MLP as base learners and combined them using an ensemble learning model with a soft voting mechanism to classify and evaluate the performance of the sitting posture dataset. MobileNet and SqueezeNet are exceptional lightweight CNN crafted primarily for handling image data, particularly RGB images. However, these models are not the best choice for one-dimensional data; traditional machine-learning methods are more suitable for this dataset.

\begin{table}[t]
    \centering
    \caption{Components and specifications of the experimental setup.}
    \begin{tabularx}{0.4\textwidth}{lX}
        \toprule
        Component & Specification  \\
        \midrule
        Camera & Azure Kinect \\
        \addlinespace
        Body Tracking SDK & V1.1.2 \\
        \addlinespace
        Desktop Depth & 72~cm \\
        \addlinespace
        Camera Elevation & 55~cm \\
        \addlinespace
        Camera Angle & 45° \update{(left \& right)} \\
        \bottomrule
    \end{tabularx}
    \label{tab:setup}
\end{table}

\update{For the SVM, we selected for a polynomial kernel function due to its ability to capture nonlinear relationships between features, thus enhancing modeling capabilities. To balance robustness and simplicity, we set the hyperparameter \( C \) to its default value of 1.0. A lower \( C \) value reduces the risk of overfitting, ensuring that the model does not become overly tailored to the training data, thereby improving its generalization on unseen datasets. This choice allows us to harness the advantages of the polynomial kernel in nonlinear modeling while maintaining computational efficiency, making it suitable for complex datasets.}

\update{For the DT, we employed the C4.5 algorithm, which not only constructs effective decision trees but also includes post-pruning techniques to minimize complexity. This optimization simplifies the model structure, leading to improved generalization performance.}

\update{In deep learning, various algorithms can handle complex data effectively. In our study, we conducted extensive comparative analysis and successfully reduced the dimensionality of posture features to 13. This reduction allowed us to use a smaller yet accurate model, particularly valuable in scenarios with limited computational resources. Given these conditions, we chose MLP, a classic deep learning architecture well-suited for non-sequential data with intricate inter-feature relationships. In our setup, the MLP has three hidden layers with configurations of 200, 100, and 25 neurons, respectively. The maximum number of iterations (\texttt{max\_iter}) was set to 200 to ensure convergence. We employed the ReLU activation function due to its well-documented performance benefits in multilayer perceptrons (MLPs). To enhance the learning process, we selected the Adam optimizer, renowned for its adaptability, which automatically manages learning rate adjustments, optimizing our model's training efficiency.}

\update{To demonstrate the effectiveness of ensemble learning algorithms, we compared our model with GBDT~\cite{NIPS2017_6449f44a} and TabNet~\cite{Arik_Pfister_2021} using a soft voting ensemble method. The \texttt{GradientBoostingClassifier}, a widely used ensemble learning model, enhances predictive accuracy by training multiple decision trees incrementally, thus reducing model loss progressively. We set \texttt{n\_estimators} to 25, striking a balance between preventing overfitting and maintaining strong performance. To ensure reproducibility, \texttt{random\_state} was fixed at 42.}

\update{For tabular data, we applied the \texttt{TabNetClassifier}, a deep learning model known for its effective feature selection and classification abilities. The model employs attention-inspired mechanisms, with \texttt{n\_steps} controlling the number of feature selection iterations, allowing for multiple perspectives on the data. Parameters \texttt{n\_d} and \texttt{n\_a} define the embedding sizes for decision and attention steps, set to 16 to balance complexity and speed. A higher \texttt{gamma} value, set to 1.5, intensifies the attention mechanism, allowing the model to emphasize important features in each step. Sparse regularization, controlled by \texttt{lambda\_sparse} (set to 1e-3), mitigates overfitting. The Adam optimizer, with a relatively high learning rate of 0.02, accelerates initial convergence. We used \texttt{mask\_type=`sparsemax'} for feature selection, promoting interpretability and reducing noise.}

\update{In particular, our} experimental setup and software configurations are shown in Table~\ref{tab:envir}. The operating system is Windows 10. The programming language used is Python, version 3.8.12, with PyCharm 2021.2.2 serving as the integrated development environment. The system is powered by an Intel(R) Core(TM) i7-10710U CPU, clocked at 1.10GHz, boosting to 1.61 GHz. An NVIDIA GeForce MX250 GPU handles graphics processing. For deep learning tasks, we utilize the Pytorch framework, version 2.0.1.

\begin{table}[t]
    \centering
    \caption{Experimental data processing environment}
    \begin{tabularx}{0.45\textwidth}{lX}
        \toprule
        Component & Specification  \\
        \midrule
        GPU & NVIDIA GeForce MX250 \\
        \addlinespace
        CPU & Intel(R) Core(TM) i7-10710U \\
        \addlinespace
        Python & V3.8.12 \\
        \addlinespace
        Operating system & Windows~10 \\
        \addlinespace
        PyCharm & 2021.2.2 \\
        \addlinespace
        Pytorch & V2.0.1 \\
        \bottomrule
    \end{tabularx}
    \label{tab:envir}
\end{table}
%%%%%%%%%%%%%%%%%%%%%%%%%%%%%%%%%%%%%%%%%%
\section{System Performance Analysis and Demonstration}
In this section, we compare the classification performance of the base learners SVM, DT, and MLP, \update{as well as the ensemble learning model using soft voting based on these base learners, alongside the comparison algorithms GBDT and TabNet on the posture dataset.} We analyze the relationship between each action through the confusion matrix by comparing the degree of misjudgment. Finally, the precision, recall, and $F_1$ scores are compared  \update{across six} algorithms. 
\subsection{Model performance evaluation indicators}
To measure the results, there are many indicators can be used to evaluate the performance of the model. Precision is the proportion of correctly predicted positives to all positive predictions. Then, there are two possibilities for prediction: one is to predict the positive class as a positive class (TP), and the other is to predict the negative class as a positive class (FP). The precise formula is as follows:
\begin{equation}
    Precision=\frac{TP}{TP+FP}
\end{equation}
Recall is the proportion of correctly predicted positive classes to all positive ones. There are two possibilities, one is to predict the positive class as a positive class (TP), and the other is to predict the positive class as a negative class (FN). The specific formula is as follows:
\begin{equation}
    Recall=\frac{TP}{TP+FN}
\end{equation}
The proportions of the six sitting and standing postures in the data set we collected are different; that is, the data set is very uneven. Hence, overall model accuracy is not an appropriate metric for measuring performance. For this, we use the $F_1$ score to evaluate the performance. The $F_1$ score takes into account both false positives (FP) and false negatives (FN) \cite{xia2020lstm}. The $F_1$ score combines precision and recall and is the harmonic mean of the two. The $F_1$ formula is as follows: 
\begin{equation}
    F_1=\sum_{i}2\cdot w_i\frac{Precision_i\cdot Recall_i}{Precision_i+Recall_i}
\end{equation}
Among them,
\begin{math}
    w_i=\frac{n_i}{N}
\end{math}, which represents the proportion of the $i$-th type of samples in the total samples. $n_i$ is the number of samples of type $i$, $N$ is the number of total samples. $Precision_i$ and $Recall_i$ are the precision rate and recall rate of the $i$-th sample. 

\begin{figure}[t]
    \centering
    \hspace*{-1cm}
    \includegraphics[width=3in]{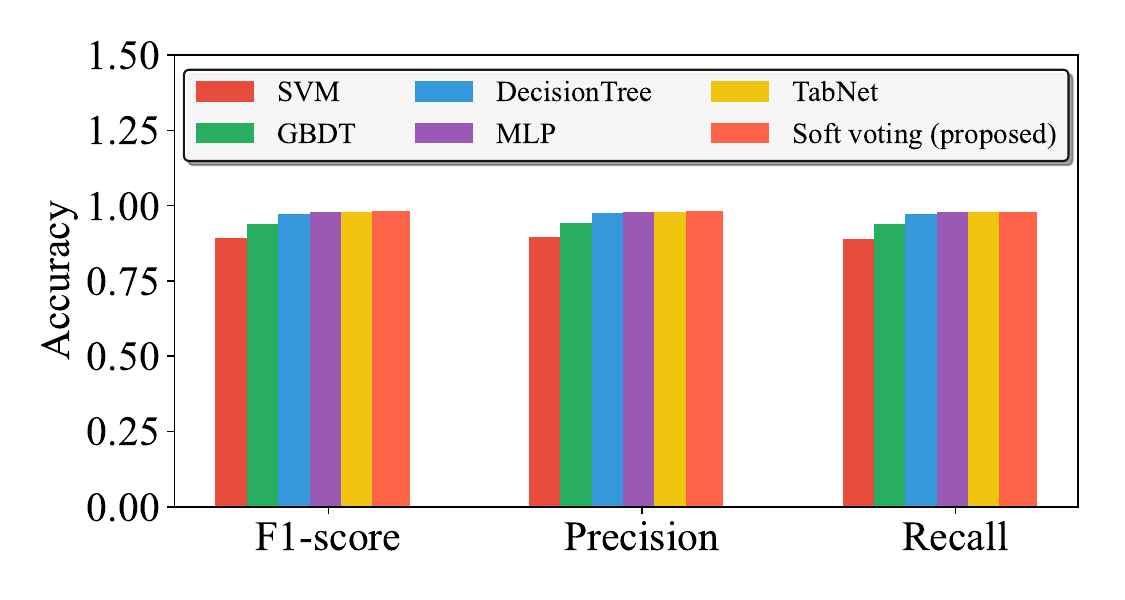}
    \caption{Recognition rate of different learning methods.}
    \label{fig:recognition}
\end{figure}
\begin{figure*}[t]
\centering
   \subfigure[MLP]{
	\begin{minipage}[t]{0.25\linewidth}
        \centering
       \includegraphics[width=1\textwidth]{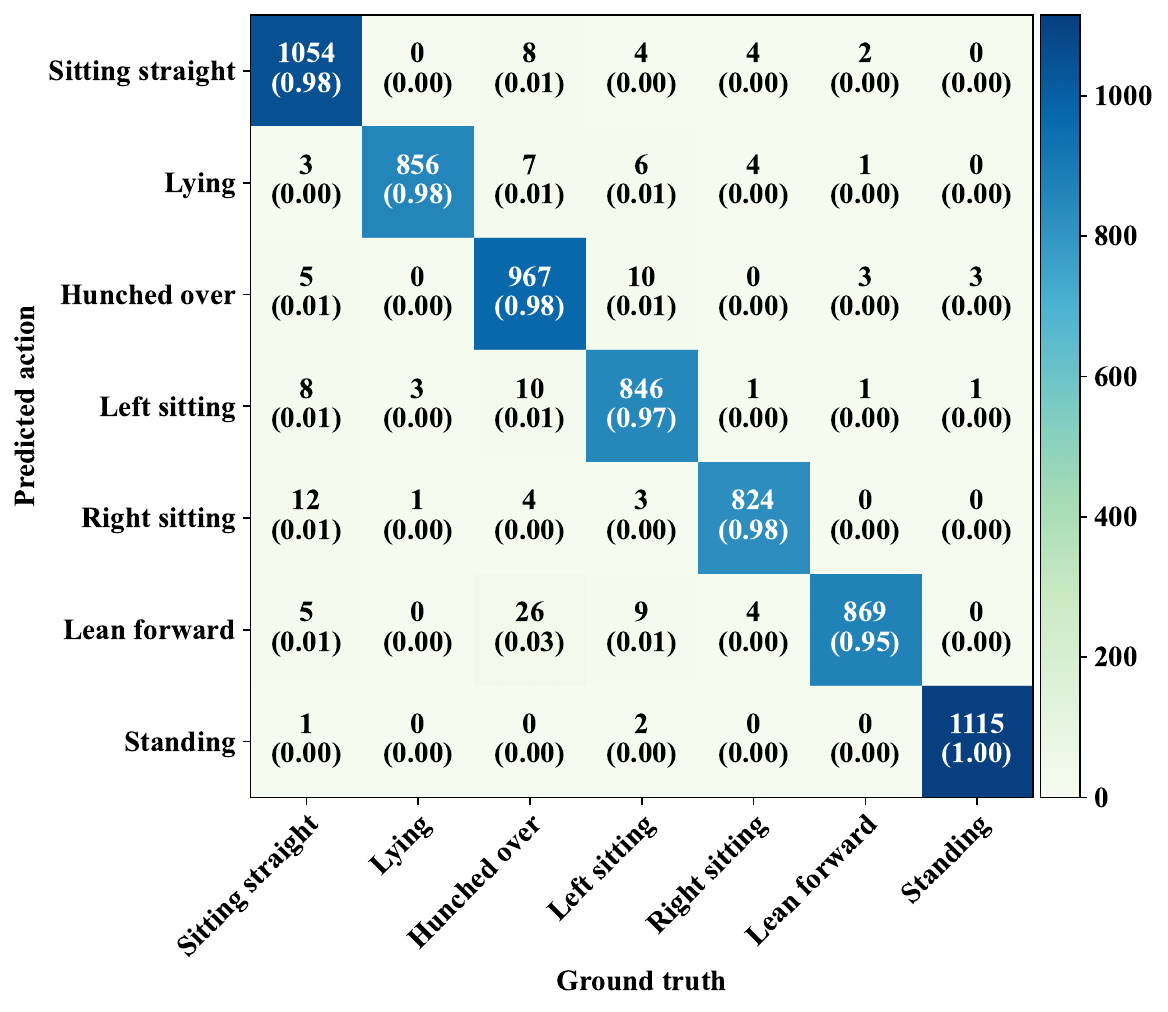}
        \label{fig:cm_cnn}
        \end{minipage}
    }
    \subfigure[SVM]{
	\begin{minipage}[t]{0.25\linewidth}
        \centering
        \includegraphics[width=1\textwidth]{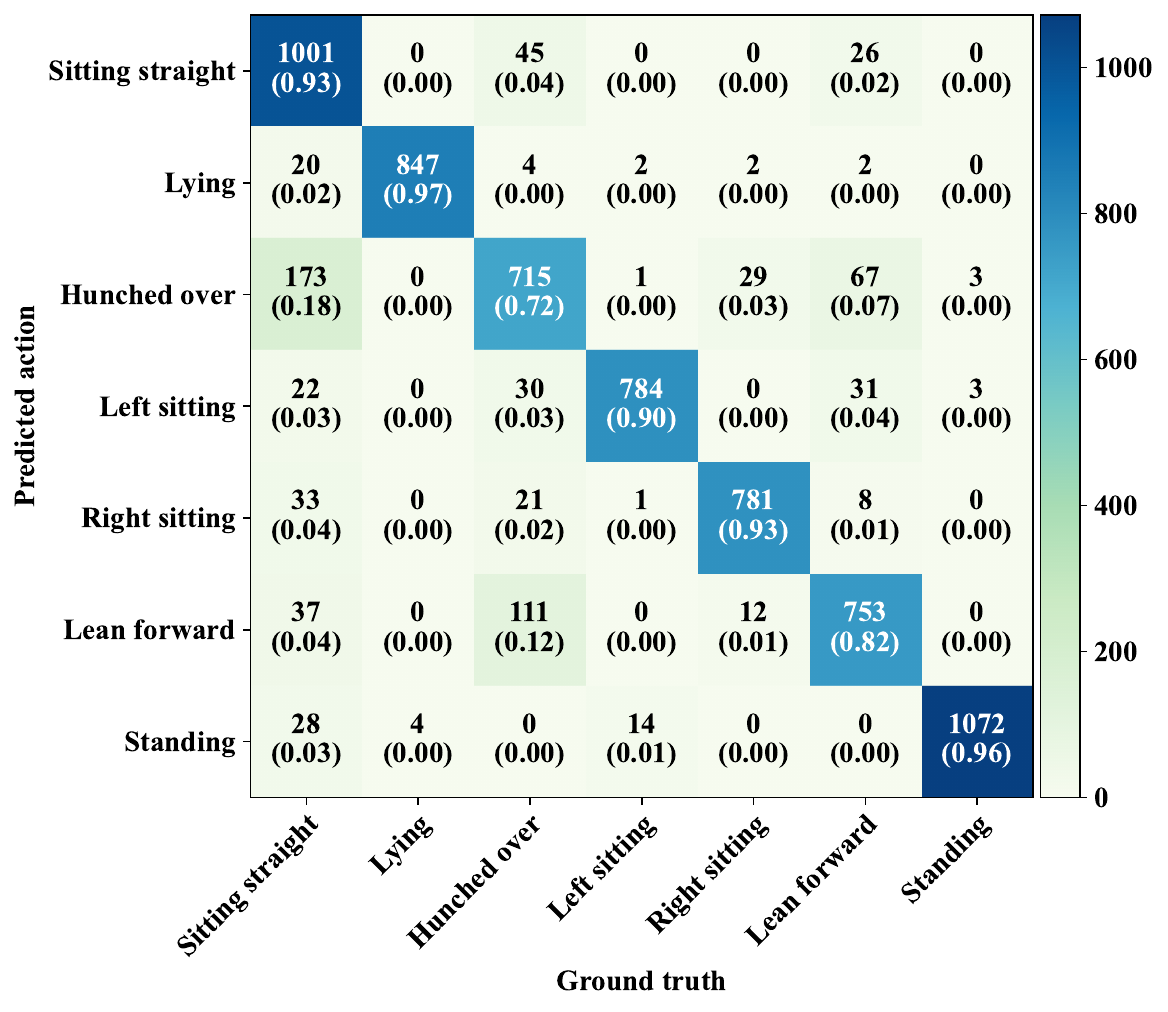}
        \label{fig:cm_svm}
        \end{minipage}
    } 
      \subfigure[Decision tree]{
	\begin{minipage}[t]{0.25\linewidth}
        \centering
        \includegraphics[width=1\textwidth]{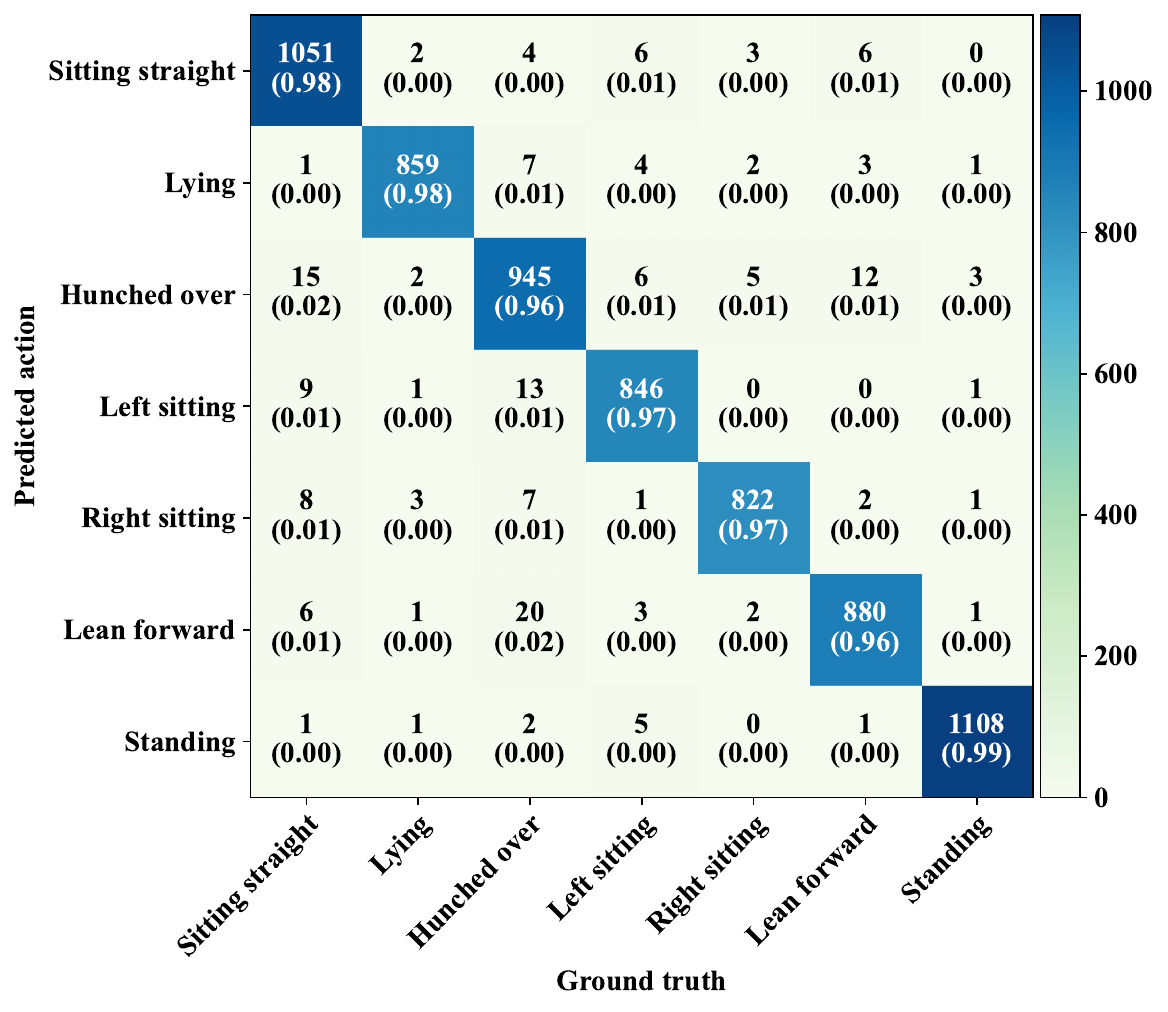}
        \label{fig:cm_decisiontree}
        \end{minipage}
    } \\
      \subfigure[GBDT]{
	\begin{minipage}[t]{0.25\linewidth}
        \centering
        \includegraphics[width=1\textwidth]{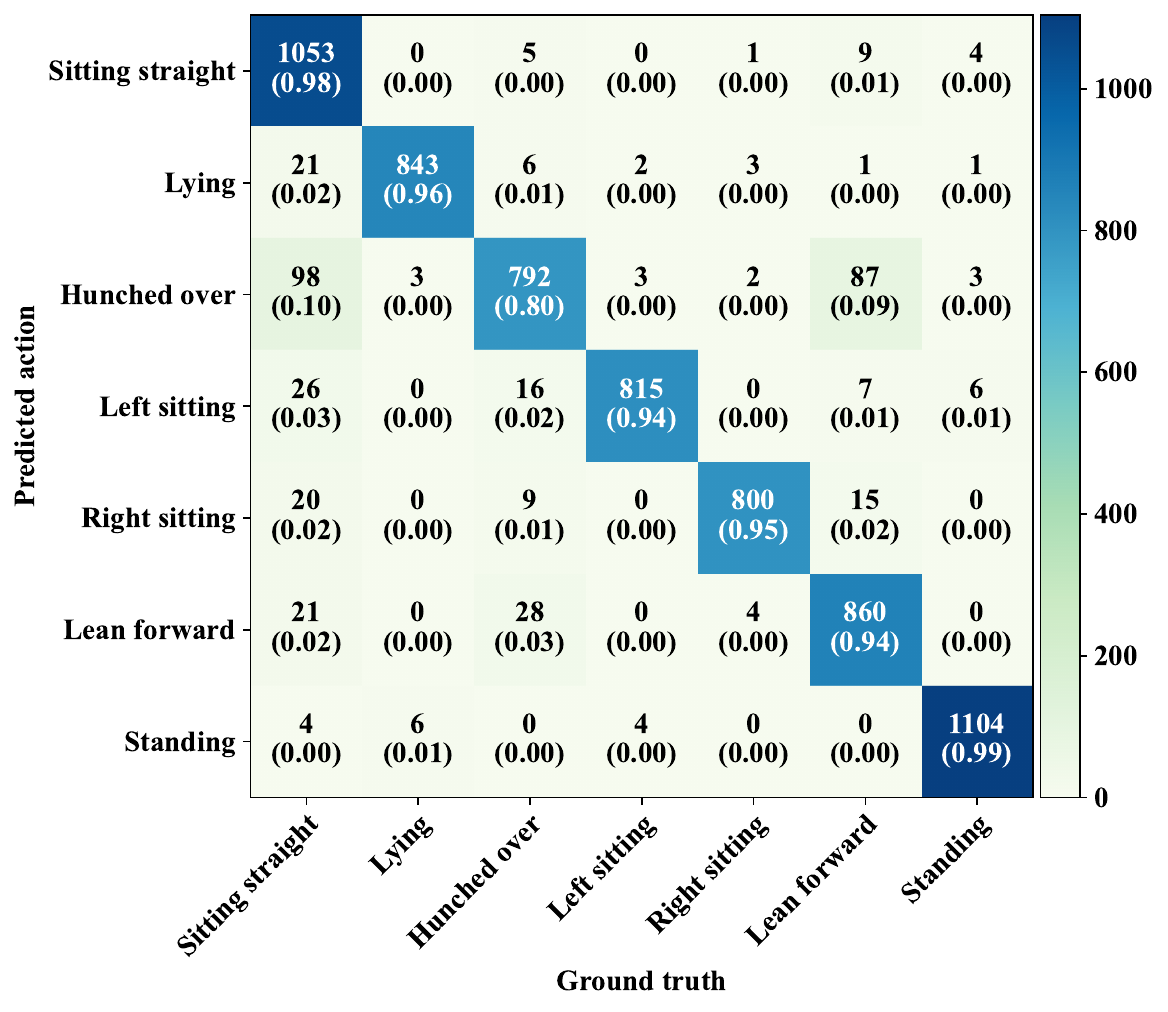}
        \label{fig:cm_decisiontree}
        \end{minipage}
    }
     \subfigure[TabNet]{
	\begin{minipage}[t]{0.25\linewidth}
        \centering
        \includegraphics[width=1\textwidth]{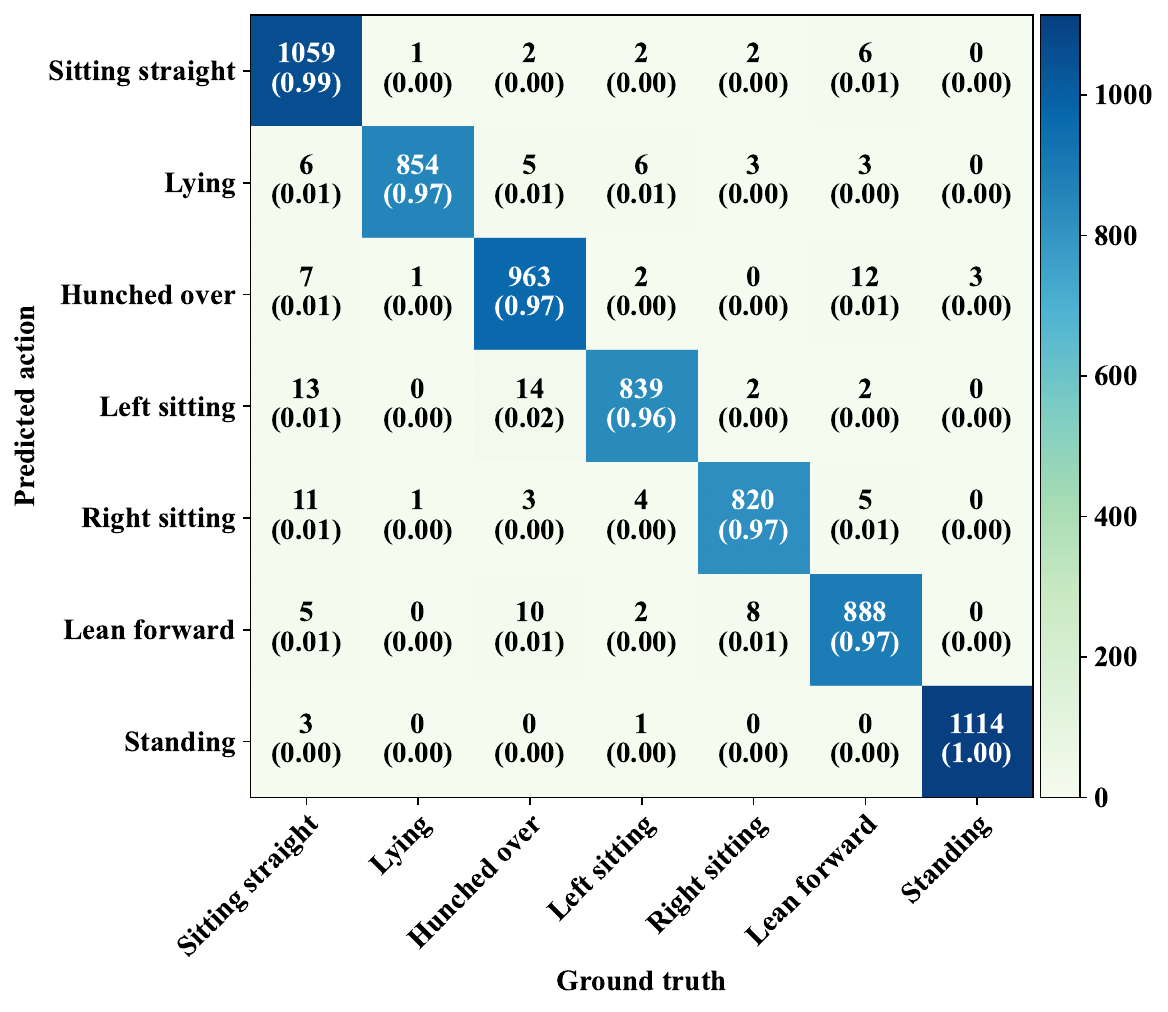}
        \label{fig:cm_decisiontree}
        \end{minipage}
    }
      \subfigure[Voting classifier \update{(proposed)}]{
	\begin{minipage}[t]{0.25\linewidth}
        \centering
        \includegraphics[width=1\textwidth]{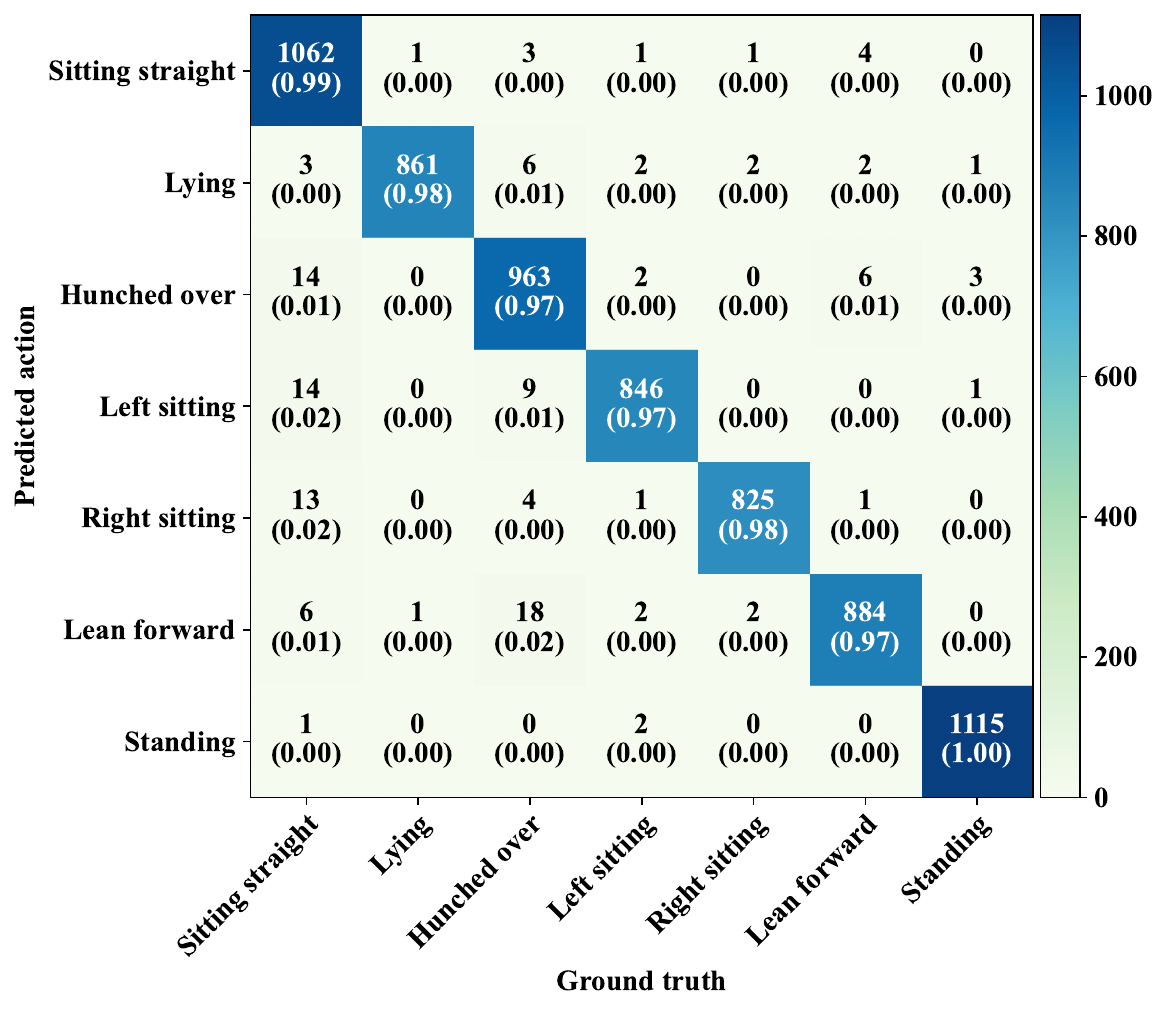}
        \label{fig:cm_naivebayes}
        \end{minipage}
    }
\caption{Confusion matrix of our considered \update{six} algorithms.}
\label{fig:confusion}
\end{figure*}

\subsection{Model Performance Comparison}
Next, we compare the precision, recall, and F1 scores of the \update{six} models. As shown in Figure 7, SVM had the lowest performance, with an F1 score of 89.1\%. GBDT performed well, achieving a score of 93.9\%. The F1 scores for the decision tree, MLP, and TabNet were all around 97\%. The ensemble learning model, which utilized a soft voting mechanism among these base classifiers, achieved the best performance with an F1 score of 98.1\%. Overall, the ensemble learning model demonstrated the most effective classification performance on our sitting dataset.

We use \update{six} artificial intelligence algorithms to recognize sitting and standing states: SVM, DT, MLP, \update{GBDT, TabNet} and an ensemble learning approach incorporating a soft voting mechanism that integrates the aforementioned base learners. To compare the accuracy of \update{these six} algorithms, \update{the obtained confusion matrix are illustrated} in Fig.~\ref{fig:confusion}. \update{Across the board, there is a notable trend where several algorithms tend to misclassify ``hunched over'' and ``leaning forward'' postures, indicating that similarities in joint angles create challenges in distinguishing these categories. For instance, SVM and Decision Tree models displayed higher misclassification rates for these similar postures, though they performed well with distinct categories like ``sitting straight'' and ``standing.''}

\update{MLP and GBDT showed better performance by capturing more nuanced patterns in the data, with GBDT particularly excelling in reducing errors for complex sitting postures such as ``left sitting'' and ``right sitting.'' TabNet leveraged its attention mechanisms to improve classification accuracy, effectively handling subtle posture variations. The ensemble learning approach, which integrated all these base learners, significantly reduced misclassification rates. By combining the strengths of each model, the ensemble method delivered the highest overall accuracy, effectively managing posture categories that posed difficulties for individual algorithms.}

\begin{figure}[t] 
    \centering
\includegraphics[width=3in]{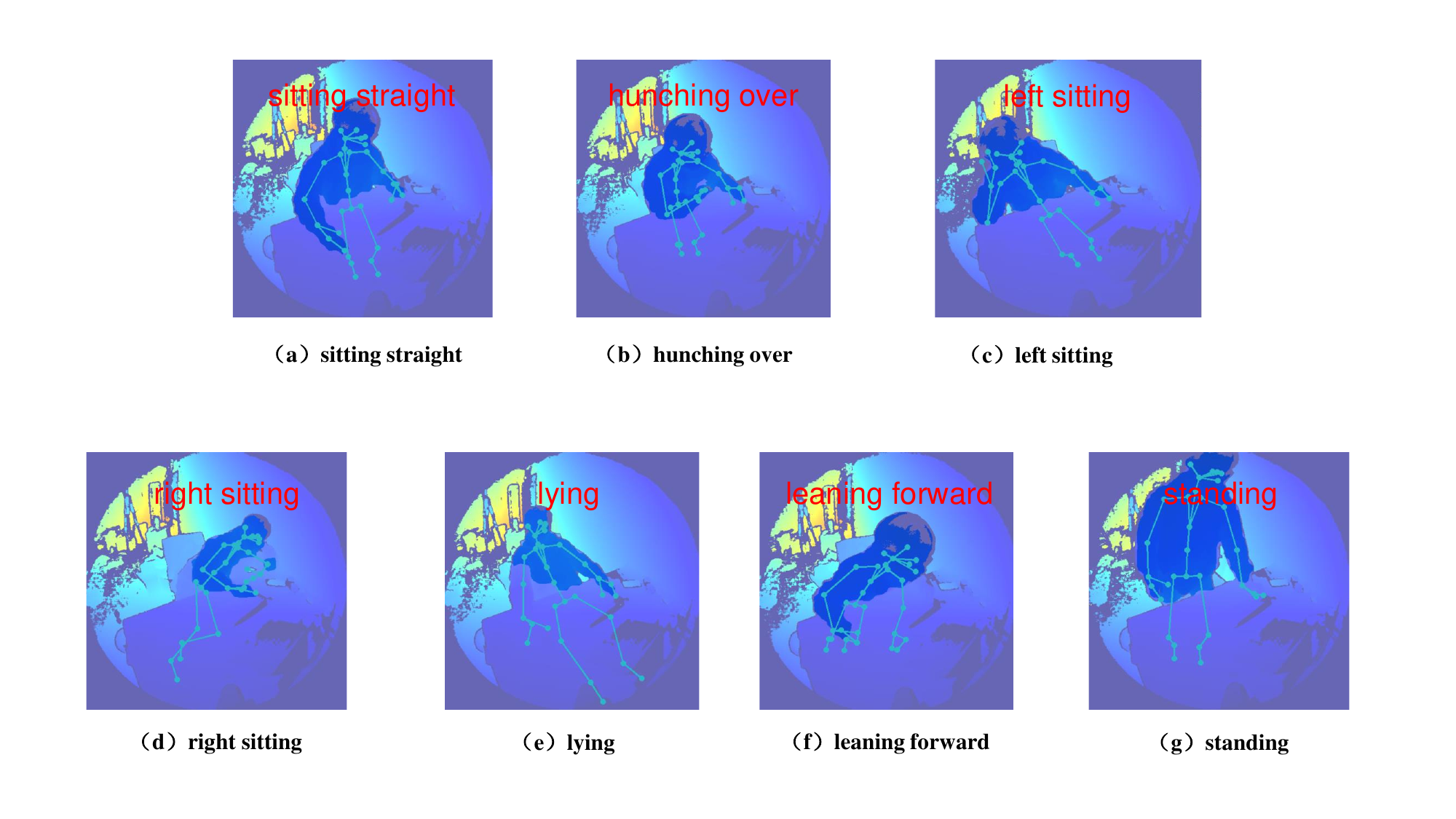}
    \caption{Skeletal depth picture for sitting position detection with detected results in our system.}
    \label{fig:skeletal}
\end{figure}
\begin{figure}[t]
    \centering
    %\hspace*{-0.5cm}
\includegraphics[width=3in]{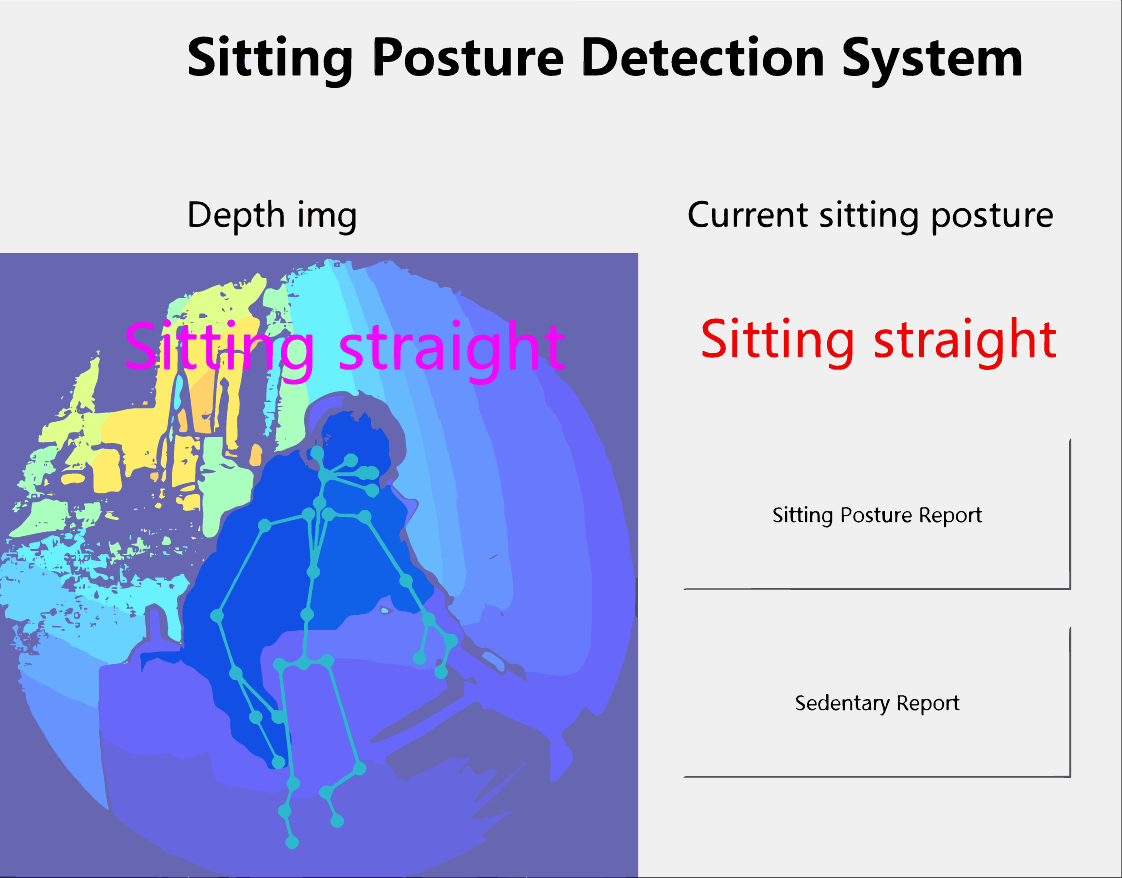}
\caption{The implemented sitting posture detection system in Windows opreating system.}
    \label{fig:sp_system}
\end{figure}

\subsection{Model Deployment and Real-time Sitting Posture Recognition}
\begin{figure*}[t]
\centering
\begin{tabular}{ccc}
\includegraphics[width=0.3\textwidth]{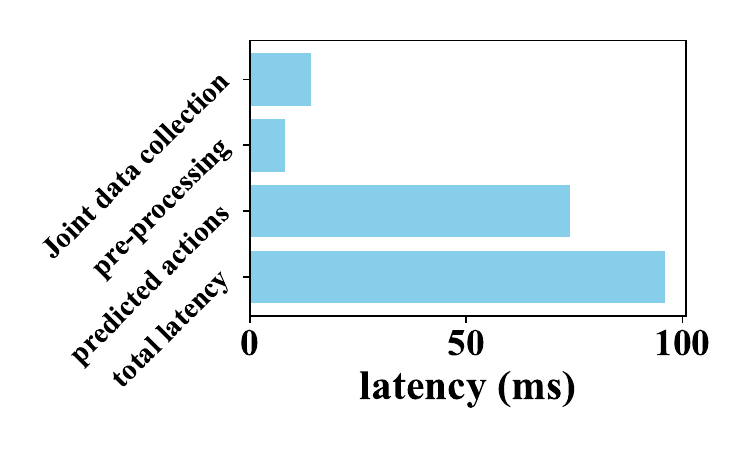} &
\includegraphics[width=0.3\textwidth]{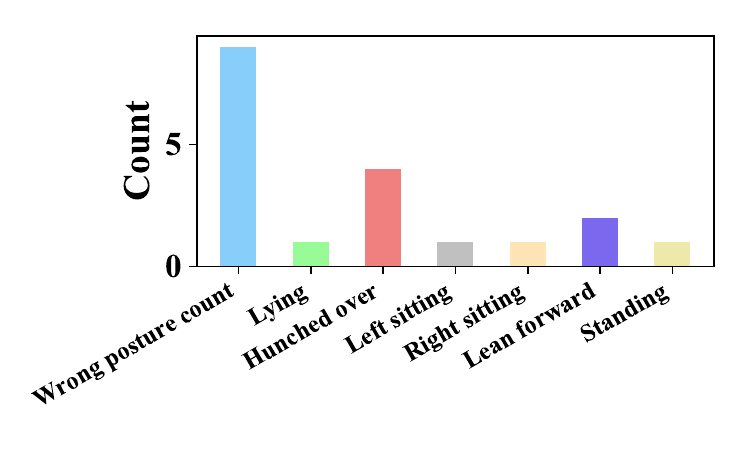} & 
\includegraphics[width=0.3\textwidth]{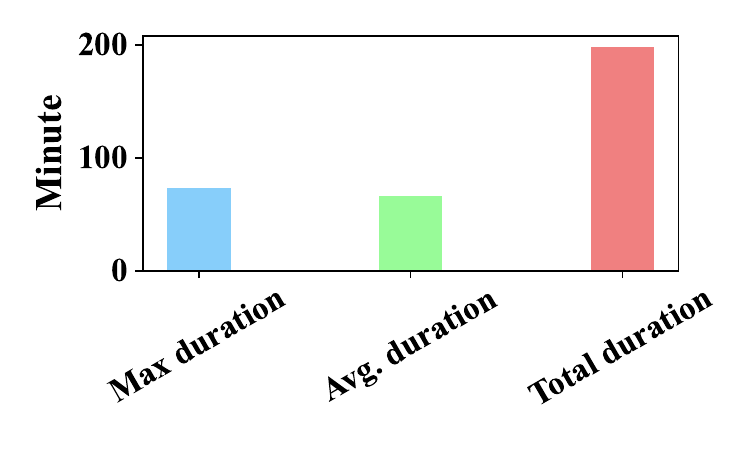} \\
(a) latency & (b) posture & (c) sedentary \\ [6pt]
\end{tabular}
\caption{The latency and generated reports of our Sitpose system.}
\label{fig:system_report}
\end{figure*}

The performance comparison reveals that the ensemble learning model with soft voting achieves the best classification results on this dataset among the four algorithms. Therefore, we have chosen to deploy this model in practice. First, we save the trained model parameters to the actual sitting posture detection system, capture the human bone joints through the Kinect real-time camera, and put the 3D coordinates of the selected \update{nine} joints along with the calculated joint angles as features into the trained classification model, and finally input the classification results to the screen in real-time. Fig.~\ref{fig:skeletal} shows the depth skeleton images of various actions acquired by the system, \update{and the implemented Sitpose system opreated in Windows opreating system is depicted in Fig.~\ref{fig:sp_system}}. Considering the recognition error and time delay, we do not judge all the classification results of the data collected every second but count the data within one minute. If the abnormal sitting posture accounts for more than 75\%, we will judge the sitting posture for this minute. If it is terrible, a pop-up window will be given to remind the users to change their sitting posture to a healthier one.

The sedentary test detects that if the users have been sitting for a cumulative hour, the system will give a pop-up window reminder to remind the users that the users are in a sedentary state and should stand up and move around to improve the sedentary problem. Real-time monitoring is essential for detecting and ameliorating poor posture habits that could culminate in spinal disorders and chronic back pain~\cite{el2011novel}. Our innovative system proffers instantaneous feedback on the user's posture, significantly diminishing the propensity for poor posture and fostering the adoption of healthier sedentary behaviors.

To assess the latency of the entire system, we conducted latency testing while using commonly used office software. In the process, Azure Kinect data was collected and underwent preprocessing and angle feature extraction, followed by category prediction and total delay time measurement. The results shown in Fig.~\ref{fig:system_report}(a) indicate that the time spent on a series of processing steps leading to the final prediction was 97 milliseconds. Furthermore, when running only the posture monitoring system without any other office software, the latency can be reduced to within 30 milliseconds. It demonstrates the real-time capability of the system.

\subsection{Testing of Sitting and Sedentary Systems}
Due to variations in the depth data and joint angle information resulting from different camera angles and positions, it is necessary to place the camera on the right side of the desk at a depth of 72 centimeters. The camera should be positioned at a downward angle of 45 degrees. This setup ensures that the entire desk area falls within the range for posture detection. We also selected two testers to participate in the test of the abnormal sitting posture and sedentary system, and these two testers did not participate in collecting the data set. With the depth camera and monitoring system turned on, we asked the tester to perform seven actions: sitting straight, hunched over, left sitting, right sitting, leaning forward, lying, and standing. 

Each pose was completed 30 times to ensure robust data collection. The quantitative results revealed that when collecting data for a normal sitting posture (sitting straight), there were 3 instances where it was misjudged as leaning forward and 2 as hunching over. In a leaning forward sitting posture, there was 4 instance where it was mistaken for a sitting straight. While in a hunching-over posture, there were 3 instances of being misjudged as sitting straight and 1 as leaning forward. No other postures were misidentified.

The system can correctly identify these sitting and standing states in real-time. When the tester sits in an abnormal posture for more than one minute, the system gives an interactive prompt in a pop-up window to remind the tester to change his current sitting posture. We let the tester sit at the workstation and work normally for the sedentary test. After working continuously for more than one hour, the system gives interactive reminder information in the form of a pop-up window in the sedentary state so that the tester gets up and moves around to avoid the sedentary state. Tests have proved that our monitoring system can monitor people's sitting posture and efficiently identify the sedentary states in real time while giving interactive prompts so that people can improve their sitting posture and avoid sedentary states. 

\subsection{Implementation of SitPose}
Figure~\ref{fig:sp_system} illustrates the \update{user interface of our Sitpose system which monitors sitting posture and sedentary in a real-time manner}. On the left side of the display is a real-time depth image. This depth image showcases a skeletal structure, tracked using Azure Kinect's built-in body tracking feature and rendered using CV2. On the right side of Fig.~\ref{fig:sp_system}, the system displays the current sitting posture status, monitored in a real-time manner. This system encompasses two key real-time features: monitoring of the sitting posture and sedentary report.

The posture report includes the wrong posture count and the frequency of five incorrect sitting and standing postures. This report helps users analyze the number and types of incorrect postures they adopted after activating the monitoring system. By examining the frequency of incorrect postures, users can understand which wrong postures they are more prone to, helping them to avoid these postures in the future, as shown in \update{Fig.~\ref{fig:system_report}(b)}.

The sedentary report tallies the total sitting time, average sitting duration, and maximum sitting duration since the Sitpose was activated. It allows users to analyze their daily sedentary patterns. When sedentary behavior is detected, the system alerts the user, suggesting they stand up or move around, shown in \update{Fig.~\ref{fig:system_report}(c)}.

Based on the above, we have developed a visual interface for the system that can display depth images and current sitting postures in real time. When users exhibit poor sitting postures or sedentary behaviors, the system will prompt a pop-up reminder and record the posture information. This information is then visually represented through bar charts. With the integration of these reporting features, users can gain a comprehensive understanding of their sitting patterns and make informed decisions to improve their health and well-being.

\section{Conclusion}
This paper adopted the Kinect depth camera to implement a sitting posture detection system, namely SitPose. Using the human body tracking SDK, SitPose identifies the coordinates of human bone joints and calculates the angles of related joint points. We then established a data set containing 3D joint coordinates of the human body and the angle between the key points, a total of \update{33,409} pieces. We designed a machine learning-based algorithm to classify the sitting postures and long-term sitting behavior in working scenarios. A comprehensive performance evaluation was conducted to validate the effectiveness of SitPose, demonstrating its accuracy. The ensemble learning model based on the soft voting mechanism achieved an impressive $F_1$ score of \update{98.1}\%, indicating outstanding performance. Finally, we deployed the trained ensemble learning model to implement the real-time SitPose system.

\update{Our study now includes a participant pool of 36 individuals. While this number provides a more substantial dataset, it may still not fully encompass the full diversity of body types and postures, potentially affecting the generalizability of the findings.} Additionally, the system`s evaluation in a controlled environment might not mirror its effectiveness in actual, dynamic workplaces. Our future research will focus on enlarging the participant group to enrich data diversity and conducting tests in varied settings to validate the system`s effectiveness beyond controlled conditions. 

Since sitting posture recognition based on the three-dimensional coordinates and angles of bones requires the distance, direction, and height of the camera, in the future, we plan to recognize sitting postures at various angles and orientations, comparing the best distance, direction, and angle for sitting posture recognition while also realizing multiple human sitting posture detection.

\bibliographystyle{IEEETran}
\bibliography{reference}

\end{document}